\documentclass[]{bytedance_seed}

\usepackage{tocloft}
\usepackage[toc,page,header]{appendix}

\usepackage{minitoc}
\usepackage{float}
\usepackage{multirow}
\usepackage{array}
\usepackage{makecell}
\usepackage{framed}
\usepackage{amsmath} 
\usepackage[colorinlistoftodos]{todonotes}
\usepackage{longtable}
\usepackage{hhline}
\usepackage{fancyvrb}
\usepackage{fvextra}
\usepackage{multicol}
\usepackage{tablefootnote}
\usepackage{threeparttable}
\usepackage{tabularx}
\usepackage{mdframed}
\usepackage[usestackEOL]{stackengine}

\usepackage{tcolorbox} 
\usepackage{xcolor}    
\usepackage{graphicx}  

\newcommand{\commentout}[1]{}

\renewcommand{\paragraph}[1]{\noindent\textbf{#1.}\hspace*{1em}}

\usepackage{enumitem}

\setlist[itemize]{leftmargin=15pt}

\RequirePackage{xspace}
\makeatletter
\DeclareRobustCommand\onedot{\futurelet\@let@token\@onedot}
\def\@onedot{\ifx\@let@token.\else.\null\fi\xspace}

\makeatother
\newcommand{\doubao}{MedXIAOHE}
\newcommand{\vit}{{Seed-ViT}\xspace}

\definecolor{myLightCyan}{HTML}{EFFDFD}
\definecolor{myTeal}{HTML}{029491}

\title{ \doubao{}: A Comprehensive Recipe for Building Medical MLLMs }

\author{ByteDance XiaoHe Medical AI}

\contribution{See \hyperref[sec:contributions]{Contributions} section for a full author list.}

\abstract{
We present \textbf{\doubao{}}, a medical vision-language foundation model designed to advance general-purpose medical understanding and reasoning in real-world clinical applications.
\doubao{} achieves state-of-the-art performance across diverse medical benchmarks and surpasses leading closed-source multimodal systems on multiple capabilities.
To achieve this, we propose an entity-aware continual pretraining framework that organizes heterogeneous medical corpora to broaden knowledge coverage and reduce long-tail gaps (e.g., rare diseases).
For medical expert-level reasoning and interaction, \doubao{} incorporates diverse medical reasoning patterns via reinforcement learning and tool-augmented agentic training, enabling multi-step diagnostic reasoning with verifiable decision traces. 
To improve reliability in real-world use, \doubao{} integrates user-preference rubrics, evidence-grounded reasoning, and low-hallucination report generation, with improved adherence to medical instructions. 
We release this report to document our practical design choices, scaling insights, and evaluation framework, hoping to inspire further research.
}

\correspondence{\email{yanghaihua@bytedance.com}}

\begin{document}
\maketitle

\begin{figure*}[!ht]
    \centering
    \includegraphics[width=0.98\linewidth]{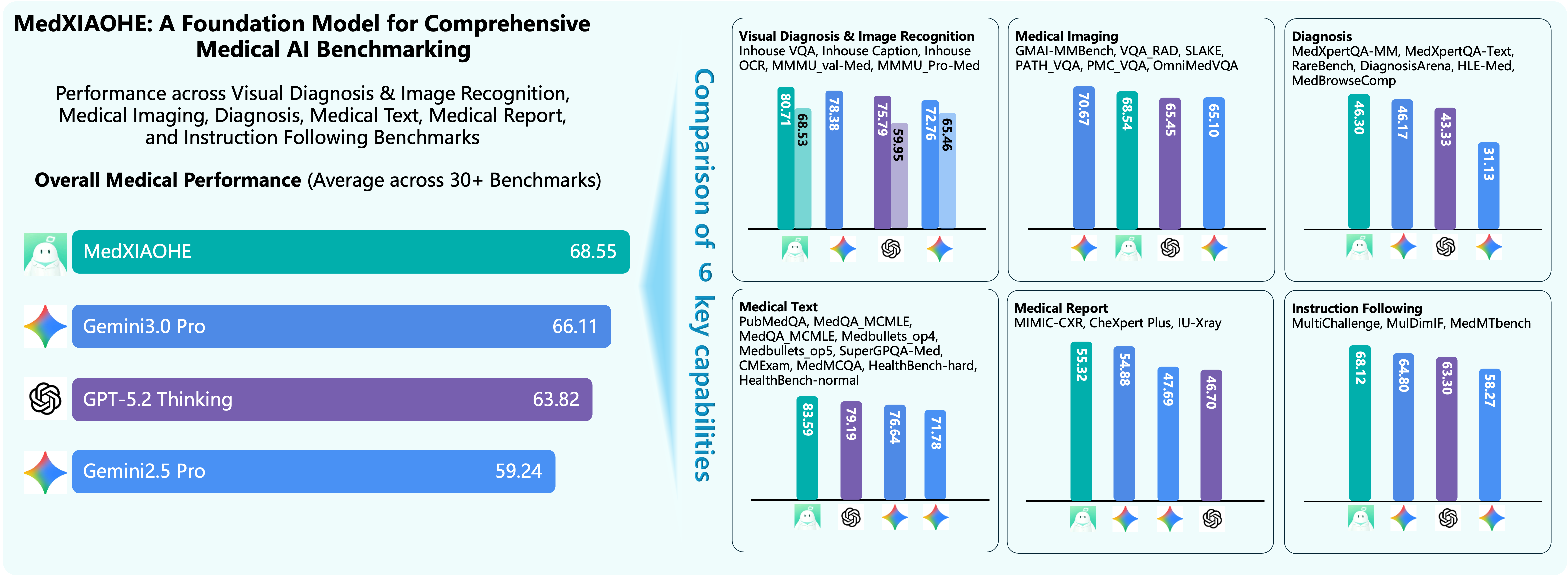}
    \caption{\textbf{Performance comparison of \doubao{} against SOTA models on comprehensive medical benchmarks.} The left panel shows the overall average score across 30+ benchmarks, demonstrating the strong performance of \doubao{}. The right panels detail the comparative results across six key capabilities. In the upper-left bar chart, dark bars represent scores on public benchmarks, and light bars represent scores on in-house benchmarks. We did not evaluate in-house benchmarks on Gemini 3.0 Pro because of changes in its privacy protocols.}
    \label{fig:model}
\end{figure*}

\newpage
\tableofcontents
\newpage

\section{Introduction}
\label{sec:introduction}

Foundation vision–language models (VLMs) have made rapid progress in recent years, showing strong ability to connect visual information with natural language. 
As these models scale, they have become capable multimodal assistants that can follow instructions and reason over complex inputs.
Healthcare is a natural and high-impact next step, given that clinical decision-making is inherently multimodal, involving patient symptoms, medical histories, physical examinations, radiology and pathology images, and structured reports. 
Emerging clinical systems, referred to as Medical Agents, leverage multimodal data and internal states to orchestrate adaptive workflows and interact with medical information systems under strict governance~\cite{hu2025landscape}.
Encouraging progress has already emerged from large medical VLM efforts \citep{medgemini2024, sellergren2025medgemma, moor2023med, llavamed2023, pan2025medvlm,chen2024huatuogptvisioninjectingmedicalvisual, lingshu2024}, suggesting that a single medical foundation model may eventually support a wide range of clinical-facing tasks.

At the same time, building a medical foundation model that is genuinely applicable in the real world remains difficult. Clinical tasks demand not only accuracy, but also disciplined evidence use, stable behavior across diverse settings, and reliable long-form generation. Many high-value scenarios sit in the medical long tail (e.g., rare diseases, atypical presentations, complex comorbidities), where coverage gaps and spurious shortcuts can surface as confident but incorrect answers. Moreover, medical inputs are heterogeneous: imaging protocols vary across institutions, visual inspection images are captured under uncontrolled conditions, and reports can contain subtle details whose omission or distortion changes clinical meaning ~\cite{Chelli2024HallucinationRates, Farquhar2024Detecting}. These characteristics make healthcare a stringent test of robustness and reliability for multimodal foundation models.

Another practical obstacle is evaluation. Medical multimodal progress is often reported across a patchwork of benchmarks with inconsistent prompting, scoring, and contamination control, which makes results hard to reproduce and hard to compare. In parallel, capabilities that strongly influence product readiness, such as medical report OCR robustness and lesion-level description faithfulness, are absent or under-measured from many public leaderboards. The broader biomedical AI community has increasingly emphasized rigorous documentation and reproducibility practices \citep{collins2021aime}, reinforcing the need for evaluation that is standardized, interpretable, and actionable for iterative development.

In this report, we propose \textbf{\doubao{}}, a medical vision-language foundation model built to narrow the gap between benchmark performance and clinical-facing usability. \doubao{} unifies medical text understanding, visual inspection, medical imaging, medical OCR, and high-quality long-form report generation within a single model. 
To scale medical knowledge while reducing long-tail gaps, we introduce entity-aware, native, multimodal continual pretraining that organizes heterogeneous corpora around an entity-centric taxonomy, thereby improving coverage and robustness across specialties and modalities. 
To improve long-form reliability, we build an evidence-grounded captioning and report pipeline that combines multi-stage drafting, medical-entity extraction and retrieval, critique-and-rewrite, and rubric-trained generative reward models to suppress subtle hallucinations and over-interpretation in clinical descriptions.
Beyond static question answering, real clinical workflows are interactive: high-quality assistance often requires gathering missing information, consulting references, and verifying intermediate hypotheses.
\doubao{} therefore integrates clinical thinking patterns with reinforcement learning and tool-augmented agentic training, enabling multi-step diagnostic reasoning with verifiable decision traces \citep{phan2025humanity, medbrowsecomp2025, curebench2025}. 
We also emphasize application-friendly behavior, including robust instruction-following and alignment with health-oriented preferences in evaluations \citep{healthbench2025}.

Another contribution of this work is a unified evaluation framework for medical VLMs. We introduce a Unified Med-VLM Benchmark that consolidates 30+ public benchmarks under a standardized protocol with unified prompting, scoring, and decontamination, enabling reproducible measurement and more meaningful cross-model comparisons. We complement public datasets with targeted inhouse evaluations designed to stress clinically important but under-measured capabilities. Together, these components provide a practical measurement backbone that connects the model and real-world medical practice.

In summary, this report provides the following highlights:
\begin{itemize}
    \item \textbf{State-of-the-Art Medical VLM Performance.} We report strong end-to-end results for medical VLMs across more than 30 public and in-house benchmarks.

    \item \textbf{Entity-Aware Continual Pretraining.}
    We propose an entity-centric continual pretraining strategy that expands medical knowledge coverage and improves robustness to long-tail cases across specialties and modalities.

    \item \textbf{Robust Reasoning Across Diverse Scenarios.}
    Our post-training approach supports reliable reasoning for text-only, vision-only, and multimodal tasks, including clinical questions that require the integration of mixed evidence.

    \item \textbf{Advanced Agentic Diagnostic Capability.}
    We demonstrate the effectiveness of tool-augmented, multi-step diagnostic reasoning with verifiable decision traces on complex, real-world diagnostic tasks.

\end{itemize}


\begin{figure*}[t]
    \centering
    \includegraphics[width=\linewidth]{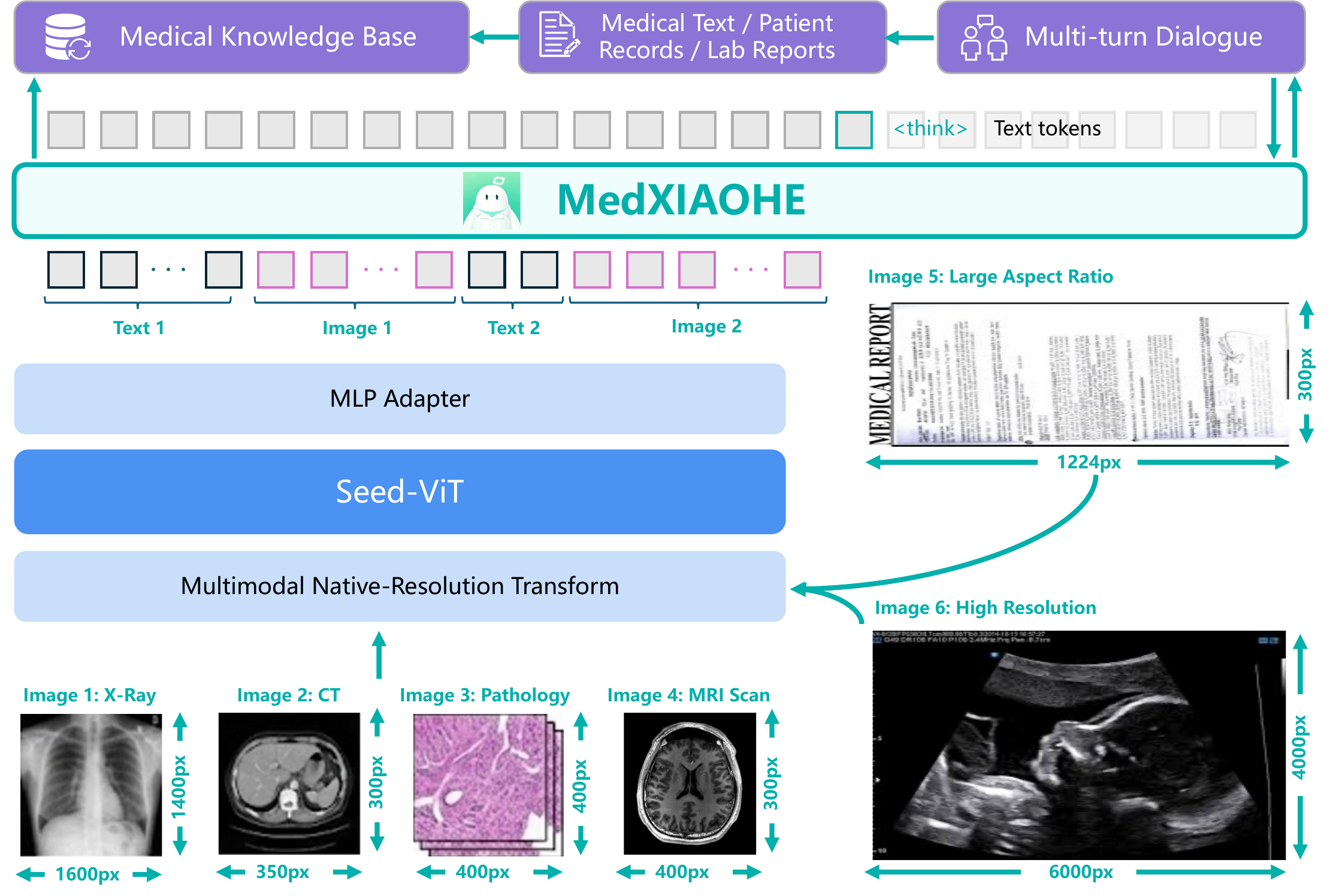}
    \caption{\textbf{The architecture of \doubao{}.} The model utilizes a Multimodal Native-Resolution Transformer to process diverse medical imaging modalities (e.g., X-ray, CT, Pathology) with varying resolutions and aspect ratios. Visual features encoded by Seed-ViT are projected via an MLP Adapter and interleaved with text tokens, integrating medical knowledge and patient records to support multi-turn dialogue and reasoning-based generation.}
    \label{fig:method}
\end{figure*}

\begin{table*}[!ht]
\caption{\textbf{Performance of \doubao{} on the unified Med-VLM benchmark compared to SOTA models.} All benchmarks are evaluated with greedy decoding. We report Pass@1 in these benchmarks. For all other benchmarks, higher numbers are better. The highest score in each benchmark is marked in bold, and the second is underlined. We did not evaluate in-house benchmarks on Gemini 3.0 Pro because of changes in its privacy protocols. }
\label{tab:medical_model_comparison}

\resizebox{1\textwidth}{!}{
\centering
\small
\setlength{\tabcolsep}{3pt}
\begin{tabular}{@{}l|l|cccc@{}}
\toprule
\textbf{Capability} & \textbf{Benchmark} & \textbf{GPT-5.2 Thinking} & \makecell{\textbf{Gemini3.0} \textbf{Pro}} & \makecell{\textbf{Gemini2.5} \textbf{Pro}} & \textbf{\doubao{}} \\
\midrule
\multirow{5}{*}{\makecell[l]{Visual Diagnosis \& \\  Image Recognition}} & Inhouse VQA & 68.27 & -- & \underline{69.64} & \textbf{76.77} \\
 & Inhouse Caption & 36.71 & -- & \underline{45.92} & \textbf{46.89} \\
 & Inhouse OCR & 74.86 & -- & \underline{80.82} & \textbf{81.92} \\
 & MMMU\_val-Med\citep{mmmu2024} & 79.73 & \underline{83.33} & 79.60 & \textbf{87.53} \\
 & MMMU\_Pro-Med\citep{mmmupro2025} & 71.85 & \underline{73.43} & 65.91 & \textbf{73.88} \\
\midrule
\multirow{6}{*}{\makecell[l]{Medical Imaging}} & GMAI-MMBench\citep{gmaimmbench2024} & 60.42 & \textbf{66.81} & \underline{61.67} & 55.21 \\
 & VQA\_RAD\citep{vqarad2018} & \underline{68.96} & \textbf{74.50} & 67.41 & 66.96 \\
 & SLAKE\citep{slake2021} & 72.83 & \underline{76.22} & 70.20 & \textbf{82.62} \\
 & PATH\_VQA\citep{pathvqa2020} & 47.88 & \underline{54.37} & 52.60  & \textbf{59.15} \\
     & PMC\_VQA\citep{pmcvqa2023} & \underline{68.65} & \textbf{70.15} & 66.85 & 63.90 \\
 & OmniMedVQA\citep{Hu2024OmniMedVQA} & 73.95 & \underline{81.96} & 71.86 & \textbf{83.40} \\
\midrule
\multirow{6}{*}{\makecell[l]{Diagnosis}} & MedXpertQA-MM\citep{Zuo2025MedXpertQA} & \underline{73.25} & \textbf{77.20} & 45.40 & 71.00 \\
 & MedXpertQA-Text\citep{Zuo2025MedXpertQA} & 55.27 & \textbf{69.47} & 41.67 & \underline{63.14} \\
 & DiagnosisArena\citep{Zhu2025DiagnosisArena} & \textbf{45.57} & 26.45 & 20.87 & \underline{42.08} \\
 & HLE-Med\citep{phan2025humanity} & 24.70 & \textbf{36.91} & 17.85 & \underline{25.77} \\
 & RareBench\citep{Chen2024RareBench} & \underline{42.19} & 41.00 & 36.01 & \textbf{46.79} \\
 & MedBrowseComp\citep{medbrowsecomp2025} & 19.00 & \underline{26.00} & 25.00 &\textbf{29.00} \\
\midrule
\multirow{10}{*}{\makecell[l]{Medical Text}} & PubMedQA\citep{pubmedqa2019} & 79.80 & \underline{80.80} & 75.60 & \textbf{86.00} \\
 & MedQA\_MCMLE\citep{pal2022medmcqa} & 91.48 & \underline{94.48} & 91.54 & \textbf{96.21} \\
 & MedQA\_USMLE\citep{medqa2020} & \underline{96.39} & 95.52 & 93.01 & \textbf{97.88} \\
 & Medbullets\_op4\citep{ChenFangSinglaDredze2025Medbullets} & 91.23 & \underline{91.88} & 82.79 & \textbf{95.78} \\
 & Medbullets\_op5\citep{ChenFangSinglaDredze2025Medbullets} & \underline{88.31} & 86.04 & 72.73 & \textbf{94.16} \\
 & SuperGPQA-Med\citep{pteam2025supergpqascalingllmevaluation} & 65.37 & \textbf{72.45} & 63.77 & \underline{71.94} \\
 & CMExam\citep{cmexam2023} & 89.12 & \underline{93.13} & 87.49 & \textbf{96.12} \\
 & MedMCQA\citep{pal2022medmcqa} & 84.77 & \underline{86.01} & 84.22 & \textbf{86.54} \\
 & HealthBench-hard\citep{healthbench2025} & \underline{42.04} & 17.70 & 16.80 & \textbf{46.10}  \\
 & HealthBench-normal\citep{healthbench2025} & \underline{63.34} & 48.40 & 49.80 & \textbf{65.20}  \\
\midrule
\multirow{3}{*}{\makecell[l]{Medical Report}} & MIMIC-CXR\citep{mimiccxr2019} & 46.44 & \underline{48.99} & 42.39 & \textbf{50.86} \\
& CheXpert Plus\citep{chexpertplus2024} & 41.85 & \underline{42.15} & 38.96 & \textbf{49.43} \\
& IU-Xray\citep{iuxray2016} & 51.80 & \textbf{73.51} & 61.73 & \underline{65.66} \\
\midrule
\multirow{3}{*}{\makecell[l]{Instruction Following}} & MultiChallenge\citep{multichallenge2024} & 60.00 & \textbf{66.80} & 56.60 & \underline{61.90} \\
& MulDimIF\citep{MulDimIF} & \underline{78.60} & 77.80 & 73.20 & \textbf{78.70} \\
 & MedMTbench\citep{yang2025medmtbench} & \underline{51.30} & 49.80 & 45.00 & \textbf{63.75} \\
\bottomrule
\end{tabular}
}
\end{table*}

\section{Overview and Architecture}
\label{sec:overview}

\doubao{} is built upon the Seed vision-language foundation model and follows the same core design: a strong vision encoder coupled with a large language model (LLM), trained as a unified multimodal decoder. Concretely, the model consists of three main components: (1) a vision backbone (\vit\citep{seed2025seed1_5vl}) that encodes one or multiple images into a sequence of visual tokens; (2) a lightweight multimodal connector that maps visual features into the LLM embedding space; and (3) an autoregressive LLM that performs instruction following, reasoning, and generation conditioned on interleaved visual and textual inputs.

As shown in Fig.~\ref{fig:method}, the architecture offers two practical advantages that are particularly important for medical applications. First, it unifies diverse medical tasks under a single interface: the same model can read long textual context (e.g., histories, guidelines, reports), interpret clinical photos and medical images, extract information from medical documents via OCR-style understanding, and generate structured or long-form outputs such as findings/impressions. Second, the decoder-centric design naturally supports multi-turn interaction: the model can incorporate prior dialogue turns, intermediate hypotheses, and retrieved evidence into a single context window, and produce responses that are consistent with the evolving conversation state.

\paragraph{Difference in Design}
\doubao{} extends the Seed foundation~\cite{seed2025seed1_5vl} through continual training rather than architecture re-design. 
This choice is intentional: for clinical-facing systems, reliability and coverage often depend more on (1) domain-specific data quality, (2) knowledge coverage in the long tail, and (3) post-training alignment for reasoning and preference, rather than on introducing complex architectural novelty. 

\paragraph{Performance}
\doubao{} demonstrates state-of-the-art performance across diverse benchmarks as shown in Tab.~\ref{tab:medical_model_comparison}.
\doubao{} is evaluated in \emph{thinking mode} against leading contemporary models, including GPT-5.2 Thinking, Gemini 3.0 Pro, and Gemini 2.5 Pro, using greedy decoding.

\begin{itemize}
  \item \textbf{Visual diagnosis \& image recognition.} \doubao{} achieves consistently strong results on both inhouse and public evaluations, outperforming SOTA models on Inhouse VQA/Caption/OCR and the MMMU medical series (MMMU\_val-Med, MMMU\_Pro-Med), indicating robust medical visual understanding and recognition.

  \item \textbf{Medical imaging.} The model exhibits clear strengths on heterogeneous radiology/pathology VQA tasks such as SLAKE, PATH\_VQA, and OmniMedVQA, suggesting effective cross-modal reasoning on clinical imagery.

  \item \textbf{Diagnosis.} \doubao{} shows competitive diagnostic reasoning performance on RareBench and MedBrowseComp and maintains strong results on DiagnosisArena and HLE-Med, indicating robust medical decision-making capability in complex and rare disease diagnosis.

  \item \textbf{Medical text.} Our model demonstrates strong medical knowledge and reasoning across PubMedQA, MedQA (MCMLE/USMLE), MedBullets, CMExam, MedMCQA, and HealthBench, reflecting solid competence in clinical QA and challenging medical reasoning.

  \item \textbf{Medical report generation.} \doubao{} performs well on MIMIC-CXR and CheXpert Plus, indicating reliable radiology report generation and clinical finding coverage. Results on IU-Xray are comparatively lower than the best baseline, suggesting remaining headroom for narrative  in certain report-generation regimes.

  \item \textbf{Instruction following.} The model is effective on multi-constraint and medically oriented instruction benchmarks (MulDimIF, MedMTbench), and remains competitive on MultiChallenge, supporting its practical usability in instruction-driven medical VLM applications.
\end{itemize}

\section{Continual Pre-training}
\label{sec:pretrain}

\begin{figure*}[ht]
    \centering
    \includegraphics[width=0.9\linewidth]{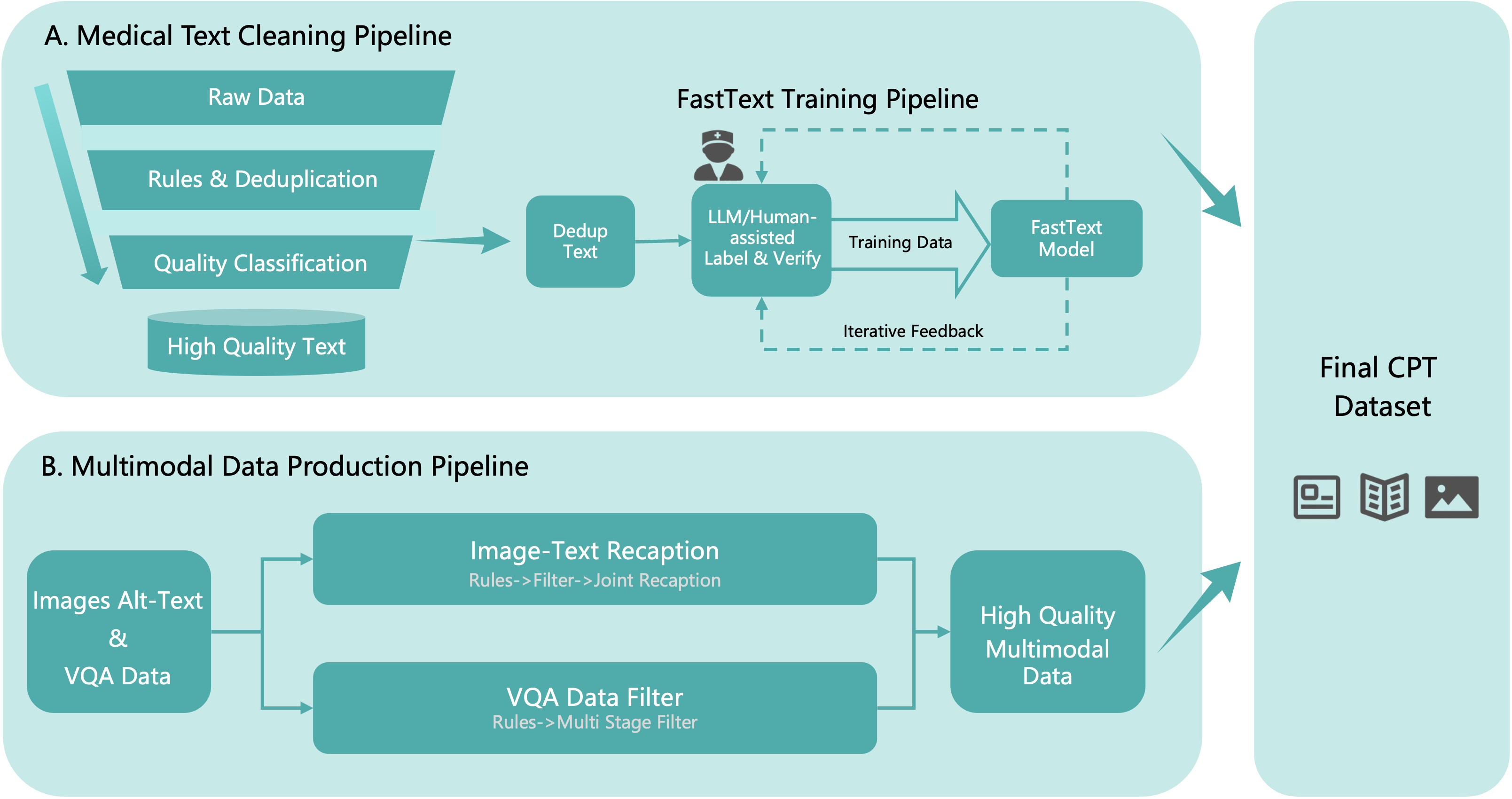}
    \caption{\textbf{Data-cleaning pipeline for the continual pretraining corpus.} The pipeline comprises two main stages: a text-cleaning workflow and a multimodal data production workflow. To construct a high-quality pretraining corpus, we apply a combination of hash-based deduplication, rule-based filtering, and model-based quality control.
}
    \label{fig:continue_pretrain_pipelie}
\end{figure*}
\subsection{Data Collection}
\label{pretrain:data_collect}
To construct a rich and factually grounded pretraining corpus, we curate data from multiple sources and apply source-specific processing to account for differences in scale and noise. In terms of volume, public web data exceeds licensed books, papers, and online clinical data; in terms of quality, the ordering is reversed.

For public web data, we collect 310B tokens and enhance using fine-grained topic classification and updated captions. Given the corpus scale and heterogeneous quality, we apply lightweight content-based deduplication, followed by model-based quality filtering, text cleaning and normalization, secondary deduplication, and image--text relevance filtering. 
For licensed medical books and papers, we perform OCR on scanned materials to obtain 280B tokens. To maximize precision, we prioritize recognition accuracy and fine-grained categorization. 
We also collect 28B tokens clinical lesion images for better visual diagnosis and 22B tokens open-source datasets. 
The continual pretraining corpus consists of a total of approximately 640 billion tokens.

Next, as shown in Fig.~\ref{fig:continue_pretrain_pipelie}, we adopt a three-stage data cleaning pipeline: 
(1) global deduplication, where we apply hash-based methods to remove redundant samples at scale; 
(2) rule-based filtering and normalization, where we use simple heuristics to filter and reshape the data (e.g., splitting long documents into manageable chunks with a sliding-window strategy); 
and (3) model-based quality filtering, where we train a medical FastText~\citep{fasttext} classifier to select high-quality samples for final training. 
We build this classifier with a self-bootstrapped positive/negative loop: we first sample instances after stages (1) and (2), label them as high or low-quality, and train an initial classifier. We then sample additional instances from the remaining pool, run inference with the classifier, and perform second-pass verification with a more capable LLM. From the verified data, we uniformly sample positive and negative instances to further refine the classifier. We repeat this loop until the classifier reaches our target accuracy.

\subsection{Medical Entity Tree (MET)}
\label{pretrain:med_entity}
The construction of the Medical Entity Tree serves three pivotal objectives: (1) Balancing Entity Training, mitigating long-tail distribution issues; (2) Quantifying Knowledge Coverage, offering a metric to evaluate the breadth of medical knowledge in pre-training data; and (3) Guiding Data Collection, identifying sparse domains for targeted acquisition. 
\begin{figure*}[ht]
    \centering
    \includegraphics[width=0.8\linewidth]{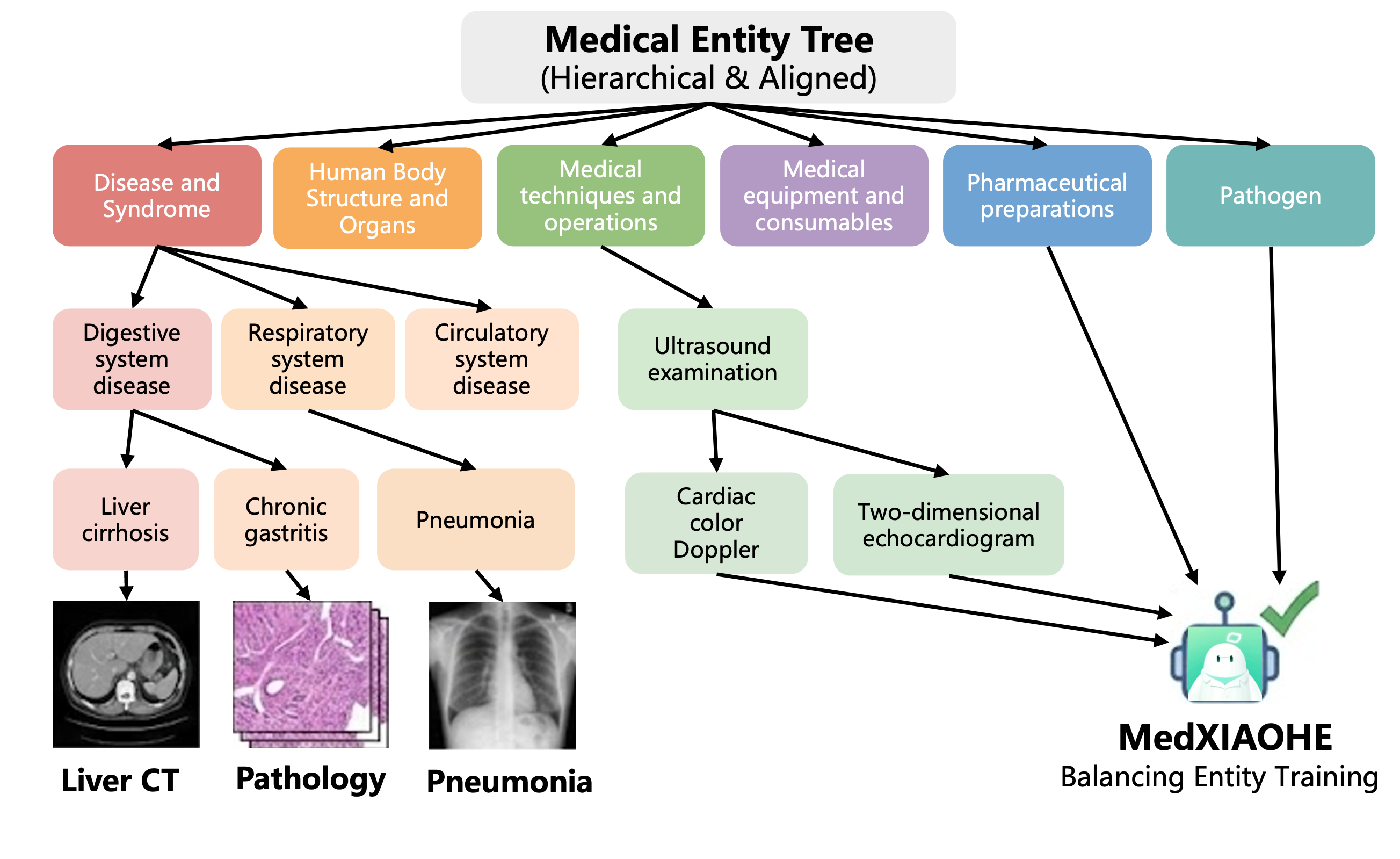}
    \caption{\textbf{Architecture overview of the Medical Entity Tree.} This hierarchical taxonomy organizes medical concepts into aligned categories to facilitate balanced entity training, precise knowledge coverage quantification, and entity-driven data acquisition.}
    \label{fig:medical_entity_tree}
\end{figure*}

\subsubsection{Medical Entity Tree Construction}

We developed a multi-stage pipeline evolving from coarse extraction to fine-grained refinement to mine entities from these materials, effectively navigating the trade-off between extraction quality and computational efficiency.
\paragraph{Stage 1: High-Efficiency Entity Extraction}
Initially, we leveraged LLMs to identify medical entities from this massive corpus. To mitigate inference latency, we implemented a \textit{Batch Chunking} strategy, identifying entities within consolidated multi-sentence input units. To ensure precision and alignment within this batched framework, we designed a rigorous prompt that enforces strict JSON formatting and domain filtering. The specific instruction is shown below:

\begin{center}
\begin{tcolorbox}[colback=myLightCyan, colframe=myTeal, title=\textbf{Prompt for Stage 1: Batch Entity Extraction}]
\small
\texttt{
Below are several sentences. Analyze them for medical entities and output according to the following requirements:
\begin{enumerate}
    \item Determine if it is a medical entity; if not, do not output.
    \item Separate entity nouns with commas; do not include duplicates.
    \item If there are no medical-related entity nouns, output "None".
    \item Output strictly in JSON format. The example format is as follows: \\
    \{\{'Sentence0': 'Entity1,Entity2,...', 'Sentence1': 'Entity1,Entity2,...', ...\}\}
\end{enumerate}
Sentences:\\
\{lines\}
}
\end{tcolorbox}
\end{center}

This prompt design significantly facilitates downstream parsing. By enforcing a strict JSON output with explicit handling of null cases (outputting ``None''), it yielded a processing speedup of approximately $30\times$ while maintaining high extraction recall and effectively filtering non-medical noise.

\paragraph{Stage 2: Joint Typing and Hierarchical Clustering}
To evolve from a flat noun list to a structured system, the second stage advanced the prompt engineering strategy to simultaneously perform entity extraction and type abstraction.  We designed the prompt to address specific challenges in medical texts, such as handling noisy characters (often from OCR or formatting) and ensuring appropriate entity granularity by splitting overly long phrases. The specific instruction is shown in Supp.~\ref{app:prompts_joint_extra}.

Post-extraction, we enforced a rigorous cleansing protocol: entities appearing fewer than $10$ times and types containing fewer than $5$ entities were filtered to reduce noise. Subsequently, we applied K-Means clustering~\cite{mcqueen1967some} on entity embeddings, utilizing the Silhouette Coefficient~\cite{rousseeuw1987silhouettes} to dynamically optimize the cluster count to define secondary types. By applying a bottom-up, frequency-weighted aggregation of embeddings, we recursively constructed a multi-level entity hierarchy.

\paragraph{Stage 3: Controlled Expansion via Tree Attachment}
Upon completing Stages 1 and 2, we obtained a concise, high-quality core taxonomy. Stage 3 focuses on scaling this repository by integrating entities extracted from the broader corpus.

\textit{Incremental Tree Attachment.}
Instead of rebuilding the tree from scratch, we adopted an Incremental Tree Attachment strategy. We treat the core taxonomy as a fixed skeleton and utilize the semantic reasoning capabilities of LLMs to graft new candidate entities onto the most appropriate nodes. To ensure the integrity of the existing structure during this expansion, we designed a strict directive prompt. The LLM acts as a taxonomic expert, mandated to identify the correct insertion path without altering existing nodes. The prompt is included in Supp.~\ref{app:prompts_entity_tree_attch}.

\textit{Optimization via Deferred Insertion.}
As the taxonomy expands, fitting the entire tree structure into the LLM context window becomes infeasible. We addressed this through a ``Deferred Insertion'' strategy: we freeze the high-quality core tree as the context anchor. For subsequent insertions, the LLM predicts the insertion path relative to this frozen core. These operations are buffered and executed in batch updates, preventing context overflow while maintaining global consistency.

\textit{Conflict Resolution via ReAct Agent.}
While the deferred insertion ensures scalability, it introduces structural ambiguity: a single entity may develop high affinity with multiple parent nodes (e.g., \textit{Kluver-Bucy syndrome} appearing under both ``Neurological Disorders'' and ``Digestive Symptoms''). This creates logical conflicts where the taxonomy degrades into a graph structure with overlapping definitions. Solely relying on the internal parametric knowledge of LLMs to resolve these conflicts often leads to hallucinations. To address this, we implemented a ReAct Agent framework~\cite{yao2022react}.

We designed a specialized prompt with rigorous logic, detailed in Supp.~\ref{app:prompts_entity_conflict_enhance}.
The transition from a passive LLM to an active agent offers three critical benefits:
\begin{itemize}
    \item Hallucination Mitigation: By mandating Retrieval-Augmented Generation (RAG), we force the model to base its arbitration on real-time, objective external evidence rather than potentially outdated or hallucinated internal weights.
    \item Explainability and Auditability: Every pruning operation generates a structured log containing the ``\texttt{<Reasoning>}'' and ``\texttt{<SearchEvidence>}''. This creates a transparent audit trail, allowing human experts to verify \textit{why} a specific classification decision was made.
    \item Dynamic Knowledge Adaptation: Since the agent connects to the live web, it can correctly classify newly discovered rare diseases or recently approved drugs that may lie beyond the pre-training data cutoff of standard LLMs.
\end{itemize}
Finally, we established a rigorous five-tier taxonomy comprising 1.4 million entities.
\subsubsection{Scalable Mapping}
After constructing the medical taxonomy, the challenge shifts to efficiently mapping the tens of millions of entries in the pre-training corpus to it. We implemented the Aho-Corasick (AC) Automaton algorithm~\cite{aho1975efficient}. By constructing a Trie tree with Fail Pointers from the finalized entity dictionary, we achieve $\mathcal{O}(N)$ time complexity for corpus scanning. This decouples processing speed from dictionary size, allowing us to rapidly calculate knowledge coverage and extract representative entities from massive datasets in approximately $20$ hours (a $3\times$ speedup over parallel LLM scanning).

\subsubsection{Quantitative Evaluation of Knowledge Coverage}
The core challenge in constructing a medical taxonomy lies in defining and measuring ``comprehensiveness''. 
Our methodology is grounded in the hypothesis that open-source medical literature encapsulates the majority of currently established medical knowledge. 
Consequently, a taxonomy distilled directly from these sources serves as the ground truth. To quantitatively validate this, we propose a semantic coverage metric to compare our taxonomy against established benchmarks.

We employ AMCS, a semantic similarity metric, to measure the extent to which a Target Set ($A$) is covered by a Reference Set ($Ref$). The metric is defined as:

\begin{equation}
\text{AMCS}(A, Ref) = \frac{1}{|A|} \sum_{i=1}^{|A|} \max_{j \in [1, |Ref|]} \text{CosineSimilarity}(\mathbf{a}_i, \mathbf{r}_j)
\end{equation}

where $\mathbf{a}_i$ and $\mathbf{r}_j$ denote the embedding vectors of entities in set $A$ and set $Ref$, respectively, generated by a pre-trained encoder (e.g., Sentence-BERT~\cite{reimers2019sentence}). This metric is asymmetric, offering two distinct interpretations:
\begin{itemize}
    \item Forward Coverage $\text{AMCS}(A, Ref)$: A high value indicates that most concepts in the target set $A$ are semantically present in our Reference taxonomy. This establishes the validity of our tree.
    \item Backward Coverage $\text{AMCS}(Ref, A)$: A lower value implies that our Reference taxonomy contains numerous concepts absent from set $A$. This establishes the comprehensive superiority of our tree.
\end{itemize}

\textbf{Experimental Setup and Baselines.}
We designate our constructed Medical Entity Tree as the Reference Set ($Ref$). We selected three distinct datasets as Target Sets ($A$) for evaluation:
\begin{enumerate}
    \item Clinical Knowledge Data: clinical capability benchmarks that measure the coverage of various aspects of the model, representing clinical practice focus.
    \item Common Crawl Medical Corpus: A large-scale general dataset representing broad medical data distribution.
    \item CMeKG (Chinese Medical Knowledge Graph)~\cite{byambasuren2019cmekg}: An authoritative open-source benchmark utilizing natural language processing on massive medical texts, referencing international standards like ICD~\cite{world2009international}.
\end{enumerate}

\textbf{Results and Analysis.}
The quantitative results are presented in Tab.~\ref{tab:coverage_results}.

\begin{table}[h]
\centering
\caption{Semantic Coverage Analysis: Forward vs. Backward AMCS}
\label{tab:coverage_results}
\begin{tabular}{lccc}
\toprule
Target Set ($A$) & Forward Coverage & Backward Coverage & Interpretation \\
& $\text{AMCS}(A, Ref)$ & $\text{AMCS}(Ref, A)$ & \\
\midrule
Clinical Knowledge Data & 0.96 & 0.68 & \textit{High inclusion, Ref is broader}\\
Common Crawl Medical Corpus & 0.95 & 0.89 & \textit{High inclusion, Ref is broader}\\
CMeKG~\cite{byambasuren2019cmekg} & 0.97 & 0.79 & \textit{Near-total inclusion}\\
\bottomrule
\end{tabular}
\end{table}

The analysis yields two critical insights:
\begin{enumerate}
    \item Near-Complete Forward Coverage: Our Medical Entity Tree achieves Forward Coverage AMCS scores exceeding $0.95$ across all datasets (0.96, 0.95, 0.97). This confirms that our tree successfully encompasses the vast majority of medical concepts found in both clinical benchmarks and established knowledge graphs like CMeKG.
    \item Significant Backward Gap: Conversely, the lower Backward Coverage AMCS scores (e.g., 0.68 for Clinical Knowledge Data, 0.79 for CMeKG) demonstrate that our Medical Entity Tree contains a substantial number of long-tail entities and fine-grained concepts not present in the baselines. This empirically supports our hypothesis that extracting directly from authoritative literature yields a more comprehensive ontology than existing collections.
\end{enumerate}

\subsection{Continual Pre-training Data}
\label{pretrain:data_syn}
Based on the balanced dataset built with the medical entity tree, we construct continual pre-training data with high medical knowledge coverage. The curation procedure for each category is described in this section.

\subsubsection{Text Knowledge}

To effectively expand the model's knowledge boundaries while mitigating reliance on raw web corpora—which are often compromised by noise and factual hallucinations—we introduce a data synthesis strategy grounded in a large-scale medical knowledge graph (KG). This approach aims to augment the model with high-quality, comprehensive domain knowledge, echoing recent efforts in leveraging structured knowledge to guide reliable synthetic data generation \cite{dedhia2025bottomup,chen2025graphgen,yang2024syncpt}. We initially constructed an authoritative medical KG by collecting raw documents from medical textbooks and academic papers. Following text segmentation and high-quality filtering, we extracted precise medical entities and relationships. These elements were integrated to form a robust medical KG, serving as the cornerstone for our subsequent data synthesis.

Leveraging this KG, we generated two categories of knowledge-centric QA data to address distinct granularity requirements:
\begin{itemize}
    \item \textbf{Atomic QA}: Designed to inject discrete knowledge points. We generated independent QA pairs based on the descriptions of individual nodes (entities) and edges (relationships) within the KG.
    \item \textbf{Aggregated QA}: Designed to enhance multi-fact synthesis and comprehension. We sampled relevant subgraphs via random walks, rephrasing the interconnected entities and relationships into coherent passages. Subsequently, an LLM was employed to generate QA pairs requiring the synthesis of multiple pieces of information from the passages.
\end{itemize}

We performed large-scale data synthesis on the two categories above. To address challenges such as ambiguity from dangling references and training inefficiency caused by data volume, we implemented rigorous filtering for non-self-contained expressions and selected high-quality samples based on Perplexity. This ensured efficient training, significantly improving the model's medical knowledge coverage and compositional generalization capabilities.

\subsubsection{Image Knowledge}
Unlike general-domain vision-language models, which rely on simple object-caption pairs, medical imagery requires a nuanced understanding of anatomical structures, pathological subtleties, and varying imaging modalities. A significant bottleneck in current research is the scarcity of large-scale, high-quality image-text pairs that go beyond sparse classification labels. To address this problem, we propose a systematic pipeline for instruction generation that synthesizes detailed captions from both massive public repositories and curated inhouse datasets. Our objective is to construct a ``dense narrative'' for every image, facilitating fine-grained cross-modal alignment and empowering the model to follow complex clinical instructions with precision.

\paragraph{Dual-Source Data Acquisition: Breadth and Depth}
To ensure both the generalizability and the clinical depth of our model, we employ a dual-source data strategy.

\textit{Public Dataset Integration for Breadth.} 
We leverage established open-source datasets (e.g., MIMIC-CXR \citep{johnson2019mimic}, CheXpert \citep{irvin2019chexpert}, and ROCO \citep{ruckert2024rocov2}) to provide a foundational layer of visual variety. These datasets offer a vast quantity of samples covering common pathologies and standard anatomical views. However, raw annotations in public datasets are often noisy or limited to binary classification tags (e.g., ``Pneumonia: Positive''). 

\textit{In-House Clinical Annotation for Depth.} 
To complement the public data, we introduce a proprietary dataset collected from partner medical institutions. This dataset is characterized by high-resolution imaging and, crucially, expert-level annotations provided by board-certified radiologists. Unlike public data, our inhouse annotations include granular details such as precise lesion localization, measurements of lesion progression, subtle textural changes in tissue, and differential diagnosis notes. This ``gold-standard'' data is pivotal for training the model on rare diseases, complex composite pathologies, and institution-specific imaging protocols that are underrepresented in open sources.

\paragraph{The Caption Construction Pipeline: From Sparse Labels to Dense Narratives}
The core of our data construction pipeline is the transformation of structured annotations into natural language instructions. A model cannot learn diagnostic reasoning from a simple label like ``Label: 1.'' It requires a narrative context. We implement a multi-stage ``Label-to-Caption-to-Instruction'' transformation pipeline.

\textit{Stage 1: Structured Attribute Extraction.}
    For every image, we aggregate all available metadata. For public data, we parse DICOM headers and tabular labels. For inhouse data, we extract structured reports including findings, impressions, and ROI (Region of Interest) coordinates. We formalize this into a structured JSON object containing key-value pairs representing anatomical location, finding type, severity, laterality, and technical quality.

\textit{Stage 2: LLM-Driven Semantic Expansion.}
    Raw attributes are often disjointed. To create fluent, coherent medical text, we employ a sophisticated prompt engineering strategy using a general-domain LLM~\citep{seed2025seed1_5vl} as a rewriting engine. We feed the structured attributes into the LLM with a specific system prompt: ``Act as a senior radiologist. Convert the following structured findings into a detailed diagnostic report description. Ensure medical terminology is accurate, describe the visual appearance of the findings''. The resulting outputs are then reviewed by human annotators. This process expands a sparse input into a rich caption.

\textit{Stage 3: Multi-Granularity Instruction Formatting.}
    To enable the model to handle various user intents, we do not simply pair the image with a single caption. Instead, we generate diverse instruction-response pairs derived from the generated caption. These include: (1)
Descriptive Instructions (e.g. ``Describe the findings in this chest X-ray.''), (2)
Diagnostic Q\&A (e.g. ``Is there evidence of pleural effusion? If so, describe its location.''), (3)
Localization Tasks (e.g. ``Identify the region responsible for the patient's shortness of breath.''), and (4)
Reasoning Prompts (e.g. ``Based on the visual evidence, what is the most likely diagnosis and what follow-up is recommended?'').

\subsubsection{Caption}
To construct a high-quality dataset for medical image captioning, we employ a multi-stage synthesis pipeline utilizing a specialized fine-tuned model. 
The generation process begins by producing an initial draft caption derived from the source image's existing alt text found in large-scale raw data. 
Subsequently, we extract key medical entities (e.g., pharmaceutical names and disease names) from the initial draft to query an internal knowledge base. The retrieved domain knowledge, combined with the initial draft, is then fed into a LLM which generates specific revision suggestions.
Based on these expert-level recommendations, the initial draft undergoes a final round of modification to produce a refined and knowledgeable caption.

Following generation, the data undergoes a rigorous filtration process to ensure clinical accuracy and descriptive rigor. 
A primary component of this phase is the rubric-based Generative Reward Model (GenRM) designed to evaluate the quality of the caption.
Given that full-scale manual inspection of the synthetic data is prohibitively expensive, we sampled a subset of data for detailed human annotation, categorizing errors related to visual omissions, incorrect identifications, and lack of rigorous phrasing. 
These annotations were converted into structured rubrics, which served as the basis for training the GenRM via SFT followed by reinforcement learning.
This model automatically scores captions based on weighted objective criteria—such as anatomical localization, morphological accuracy, and the strictness of evidence-based assertions—effectively filtering out hallucinations or ungrounded diagnostic claims.

\subsubsection{OCR}

\textbf{Data Source and Collection}. Medical images with text were selected from our medical database and processed through a standardized preprocessing pipeline, including image deduplication, low-resolution filtering, aspect-ratio filtering, and corrupted image removal. To selectively recall medical-report–type images with OCR-relevant content from the corpus, we adopted a two-stage weakly supervised strategy: (1) a vision–language model~\citep{seed2025seed1_5vl}  was used to generate pseudo-labels for 10K sampled images, where “medical report” was defined as one target category; (2) these pseudo-labeled samples were then used to train a ViT-Base classifier, which was applied to the full database to retrieve approximately 9 million medical report images.

\textbf{Data Distillation, Filtering, and Augmentation.} For the recalled images, Seed1.5-VL~\citep{seed2025seed1_5vl} was further employed to distill OCR-style textual descriptions, yielding paired image–text supervision. We then cleaned the distilled data using a rule-based filtering strategy that removed approximately 0.01\% of anomalous, noisy samples. A subset of the remaining samples was subsequently subjected to human review to ensure quality. Finally, a lightweight data augmentation strategy was applied, where 10\% of the images were randomly rotated, to improve robustness to orientation variations commonly observed in real-world medical documents.

\subsubsection{Grounding}
Precise object grounding serves as the critical bridge connecting textual semantics with visual regions, a capability indispensable for verifiable medical diagnostics \cite{luo-etal-2025-vividmed}. To enhance the fundamental visual grounding precision of \doubao{}, we developed an automated pipeline to generate high-quality training data from a large-scale inhouse medical image-text dataset. 
This process begins by utilizing a prior internal model to caption images, followed by an MLLM-based visual verifier to perform rigorous, image-grounded entity extraction. Governed by a strict system prompt, the verifier cross-references textual mentions with visual evidence, extracting only verifiably present entities while filtering out abstract concepts. This establishes a robust foundation for object-level recognition.

\subsubsection{Human-in-the-loop Data Generation}
For other types of data, we develop a human-in-the-loop pipeline that is systematically incorporated into the training process.

\textbf{Label-based annotation framework.}
We adopt a standardized label-based annotation paradigm to enable scalable execution and reliable quality control across both data annotation and verification stages. 
Licensed physicians are recruited to participate in the image annotation workflow, which includes image filtering, label assignment, and result verification. 
First, annotators assess the quality of each image and discard low-quality or non-medical images. 
Next, a senior medical expert panel defines standardized annotation guidelines, covering anatomical regions or organs, lesion types, lesion attributes, disease categories, and treatment or medication recommendations. Finally, all annotation results are reviewed and corrected by the senior expert panel to ensure consistency, achieving entity-level precision and recall exceeding 95\% after expert auditing.

\textbf{VQA synthesis from annotated labels.} 
Using the curated annotated labels together with the image as inputs, we prompt LLMs to automatically synthesize VQA instructions. 
Owing to their stronger instruction-following capabilities, we employ LLMs rather than VLMs for the synthesis stage. In addition, we incorporate rewriting techniques to further increase the complexity of the generated VQA instructions.

\textbf{General VLM-based data filtering.}
We apply a two-dimensional filtering process using a general-purpose VLM, focusing on both relevance and plausibility of the synthesized VQA data. This step aims to remove extreme or abnormal cases, such as answers that are irrelevant, hallucinated, template-based, or not grounded in the image content. Each sample is evaluated and categorized along two axes—for example, image relevance (relevant / uncertain / irrelevant) and answer plausibility (plausible / uncertain / implausible). This process substantially reduces the proportion of superficially valid but practically useless VQA pairs.

\textbf{Expert VLM rollout-based sampling and data validation.}
A medical expert, VLM, is trained on the filtered VQA data and then used for rollout-based sampling to further enforce medical correctness. For each question, the expert VLM performs pass@$k$ rollouts, generating multiple independent answers that are compared against the ground-truth annotations to assess consistency. Samples for which $k$ rollouts are fully or partially correct are retained for training, while samples with all incorrect outputs undergo additional sampling or physician manual review before a final decision is made. This stage specifically targets the correction of subtle medical inaccuracies that may pass general-purpose filtering.

\subsection{Training Recipe}
\label{pretrain:train_recipe}

Optimizing heterogeneous capabilities within a single continual pre-training (CPT) process can induce gradient interference, particularly when medical specialization and general-purpose multimodal reasoning are learned concurrently~\cite{ye2024continual,guo2025revisit,song2025alleviate}. Rather than separating training into multiple phases, we adopt a single-stage recipe that relies on \emph{data ordering} to expose the model to structured curricula over a unified corpus. The central hypothesis is that the conventional random shuffling paradigm obscures informative structure in multimodal medical data, including domain, modality, and task difficulty, thereby interleaving unrelated objectives and increasing gradient variance and conflicts during optimization. In contrast, an ordered presentation of semantically coherent samples encourages locally consistent updates and yields a more stable trajectory for joint capability acquisition.

Our data ordering paradigm reorganizes the entire CPT corpus into a curriculum over \emph{groups} of examples instead of isolated instances. We first train a lightweight warm-up model on a small random subset of the data (e.g., 10\%) and compute a fixed-length embedding for every training sample by extracting token representations from an intermediate decoder layer (for efficiency) followed by token-wise average pooling. Using these embeddings, we perform dimensionality reduction with UMAP~\cite{mcinnes2018umap} and density-based clustering with HDBSCAN~\cite{mcinnes2017accelerated} to obtain semantically coherent clusters spanning medical images, reports, captions, OCR-derived text, and general-domain QA. To induce an easy-to-hard curriculum, we rank clusters using a compactness score, defined as the mean pairwise Euclidean distance between embeddings of randomly sampled instances within each cluster. We treat more compact clusters (i.e., lower scores) as easier, as they exhibit higher intra-cluster similarity and are expected to incur less gradient conflict. Clusters are then ordered in ascending compactness score.

To avoid sharp optimization shifts at cluster boundaries, we introduce smooth transitions between consecutive clusters via mixed regions that jointly sample from both clusters and randomly shuffle the samples. Each mixed region additionally includes a small replay buffer drawn from earlier clusters, which reduces forgetting while maintaining forward progress. This single-stage, ordered training schedule produces a unified pre-training process that preserves the benefits of joint optimization while systematically reducing gradient conflict through structured exposure, thereby improving stability and data efficiency without requiring explicit stage-wise parameter freezing or capability-specific phases.

\section{Mid-Training}

While foundational pre-training provides broad general knowledge, it does not fully meet the complex requirements of professional clinical diagnosis. To address this limitation, the mid-training phase is designed as a critical stage that systematically strengthens the model’s advanced reasoning abilities.
Specifically, this phase aims to: (1) develop atomic combinational skills, allowing the model to integrate individual capabilities, such as visual grounding and tool use, into coherent workflows; (2) substantially improve medical reasoning by moving from simple information retrieval to rigorous, multi-step clinical inference; and (3) build a high-quality supervision signal that offers a strong initialization for reinforcement learning, supplying reliable policy priors for later alignment and the development of more agentic behavior.

\subsection{Mid-Training Data Construction}

\begin{figure*}[h]
\centering
\includegraphics[width=1.0\linewidth]{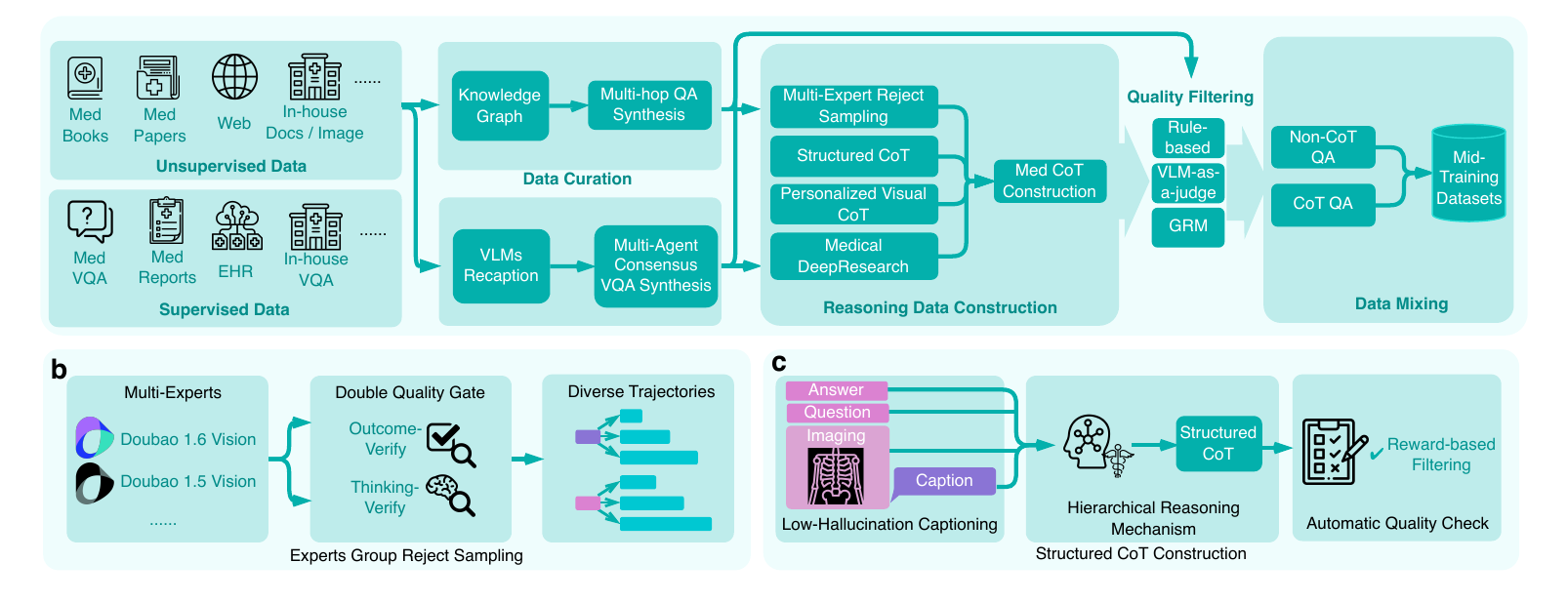}
\caption{\textbf{Mid-training data construction overview.} The framework illustrates the comprehensive pipeline designed to synthesize high-fidelity medical reasoning data from diverse sources. \textbf{a,} The data synthesis engine aggregates unsupervised and supervised corpora, utilizing knowledge graphs and multi-agent consensus to construct structured reasoning datasets. \textbf{b,} A multi-expert reject sampling mechanism with dual-quality gates is employed to distill diverse and causally valid reasoning trajectories. \textbf{c,} The process incorporates a structured Chain-of-Thought construction pipeline with automatic quality checks, strictly aligning visual perception with logical deduction to eliminate hallucinations.}
\label{fig:mid_training_overview}
\end{figure*}

A major challenge in developing medical reasoning models is the limited availability of high-quality chain-of-thought data. Although large medical corpora are available, they mainly provide final diagnoses and rarely include the explicit step-by-step reasoning needed to learn clinical logic. To address this gap, we developed a multi-dimensional data synthesis engine (see Fig.~\ref{fig:mid_training_overview}a). The framework supports diverse task formats, including multi-turn question answering, detailed image captioning, and precise visual grounding. Rather than relying on simple data augmentation, it systematically reconstructs the clinical reasoning space. By synthesizing a range of reasoning structures, from basic logical inference to complex tool-assisted decision-making, the engine provides high-fidelity training signals that better reflect the cognitive processes used by clinical experts.

\subsubsection{Internal Reasoning}

\paragraph{KG-Guided QA Synthesis}
The medical domain is characterized by complex clinical problems and diagnostic scenarios that demand robust reasoning capabilities to analyze symptoms, integrate disparate information, and derive reliable conclusions. To elevate the model's performance in this regard, we designed several data synthesis methods yielding high-quality medical reasoning data.

To enrich the diversity and complexity of medical reasoning QA data, we leverage the KG to synthesize multi-hop QA pairs. We first evaluate the model's performance on KG entities using the pass@$k$ metric to identify weak entities. For these targets, we retrieve supplementary information via our retrieval system, extracting new entities and relationships to update the KG. Starting from these weak entities, we sample complex paths (exceeding 5 hops) via random walks. The terminal node is designated as the answer, while some intermediate attributes are masked or obfuscated along the path to increase difficulty, ultimately generating complex multi-hop QA pairs and reasoning trajectories.

To tackle more complex problems and ensure the reliability and verifiability of the reasoning trajectories, we constructed a KG-based verifiable reasoning synthesis pipeline. For an existing QA pair, we first extract entities from both the question and the answer and map them to nodes in the graph through multi-stage matching. Next, we search for potential multi-hop paths between the question entities and answer entities and re-rank the candidate paths to retain those that are relevant and correct. Finally, using these filtered reliable paths as conditions, we guide the model to generate verifiable reasoning trajectories.
Notably, in the synthesis processes described above, we preserved the native reasoning patterns of our base model. This ensures the training focus remains on the reasoning content itself, avoiding performance degradation potentially caused by pattern distribution shifts.

\paragraph{Multi-Expert Reject Sampling}
Complementing the structured knowledge synthesis with broader cognitive diversity, we engineered a multi-expert reject sampling pipeline that incorporates multi-domain data (spanning visual examination, medical imaging, diagnosis, and text). This pipeline shifts synthesis from single-source generation to an ensemble-driven distillation process \cite{wang2023selfconsistency} (See Fig.~\ref{fig:mid_training_overview}b). By orchestrating expert models with varying cognitive configurations and employing best-of-N rejection sampling, we synthesize a diverse spectrum of reasoning trajectories—from concise perceptual inferences to multi-step diagnostic deductions—thereby creating a high-fidelity search space essential for generalization. We strictly enforce a dual-quality gate mechanism that systematically validates both the terminal outcome and the intermediate CoT against clinical causality. This includes Outcome-Verify—a comprehensive verification mechanism that supports judgment, selection, fill-in-the-blank, Q\&A, and other question types to ensure answer accuracy and Thinking-Verify—a mechanism that concurrently performs logical verification of the reasoning chain to ensure its soundness. Through this pipeline, we are able to build fundamental medical reasoning capabilities from scratch.

\paragraph{Structured CoT}
Complex reasoning ability is crucial for solving difficult medical problems, which often relies on high-quality Chain-of-Thought \cite{wei2022chain, zhang2023multimodal, liu2023llavacot} data. Standard forward rejection sampling often fails in complex tasks (e.g., diagnostic deduction, report generation) due to open-ended divergence.
We resolve this via a specialized Reverse Structured CoT Synthesis framework. Depending on the input modality, this approach anchors reasoning on a low-hallucination caption solely for image-based tasks to prevent perceptual errors, whereas pure text inputs proceed directly to the synthesis phase. 
To simulate the authentic cognitive process of real clinicians, we employ a medical expert role-play strategy that strictly operates without hindsight, ensuring the reasoning path is constructed without premature reliance on the ground-truth answer. 
By enforcing a four-stage structural constraint (i.e., Understanding, Visual Observation / Knowledge Recall, Reasoning and Conclusion), we align the internal rationale with formal diagnostic workflows. In the observation stage specifically, the mechanism performs visual observation for imaging data, while for textual queries, it focuses on knowledge recall to bridge current context with prior medical knowledge. Finally, an automatic quality check validates the generated rationale across six comprehensive dimensions: hindsight-free logic, logical soundness, instruction following, answer consistency, authentic thinking, and global consistency. This ensures the output is not merely a linguistic expansion but a verifiable reproduction of expert clinical derivation, effectively eliminating reasoning drift, as in Fig.~\ref{fig:mid_training_overview}c.

\paragraph{Personalized Visual CoT}
Aggressive training on long reasoning tasks can paradoxically degrade visual pattern recognition. To address this, we developed a specialized Dual-Track CoT paradigm. Our initial experiments confirmed a pervasive industry-wide bottleneck known as the perception-reasoning conflict: complex, text-heavy reasoning chains tend to detach from visual evidence, causing the model to operate purely in the textual space and leading to significant regressions in fine-grained perceptual tasks. Positing that brevity preserves fidelity—where concise reasoning prevents the dilution of visual signals and sustains visual attention—we identified that short CoT is inherently superior for perception-intensive tasks. Consequently, we developed a Personalized CoT Closed-Loop Pipeline that fundamentally reimagines the synthesis process by strictly separating logical content from cognitive style. It begins by generating a low-hallucination caption that isolates core visual details, followed by a hierarchical reasoning mechanism that constructs a first-principles logical core strictly bound to visual observations. Crucially, we employ a multi-agent architecture—comprising image analysts, reasoning experts, and style transfer specialists—to adapt this core logic into task-specific formats. By enforcing a dual-layer quality gate that validates both the structural logic and the post-adaptation style consistency, this pipeline produces high-fidelity training data that enhances reasoning depth without compromising the model’s fundamental sensitivity to visual stimuli, successfully breaking the stagnation observed in standard one-size-fits-all reasoning paradigms.

\subsubsection{Agentic Reasoning}

\begin{figure*}[!h]
    \centering
    \includegraphics[width=1.0\linewidth]{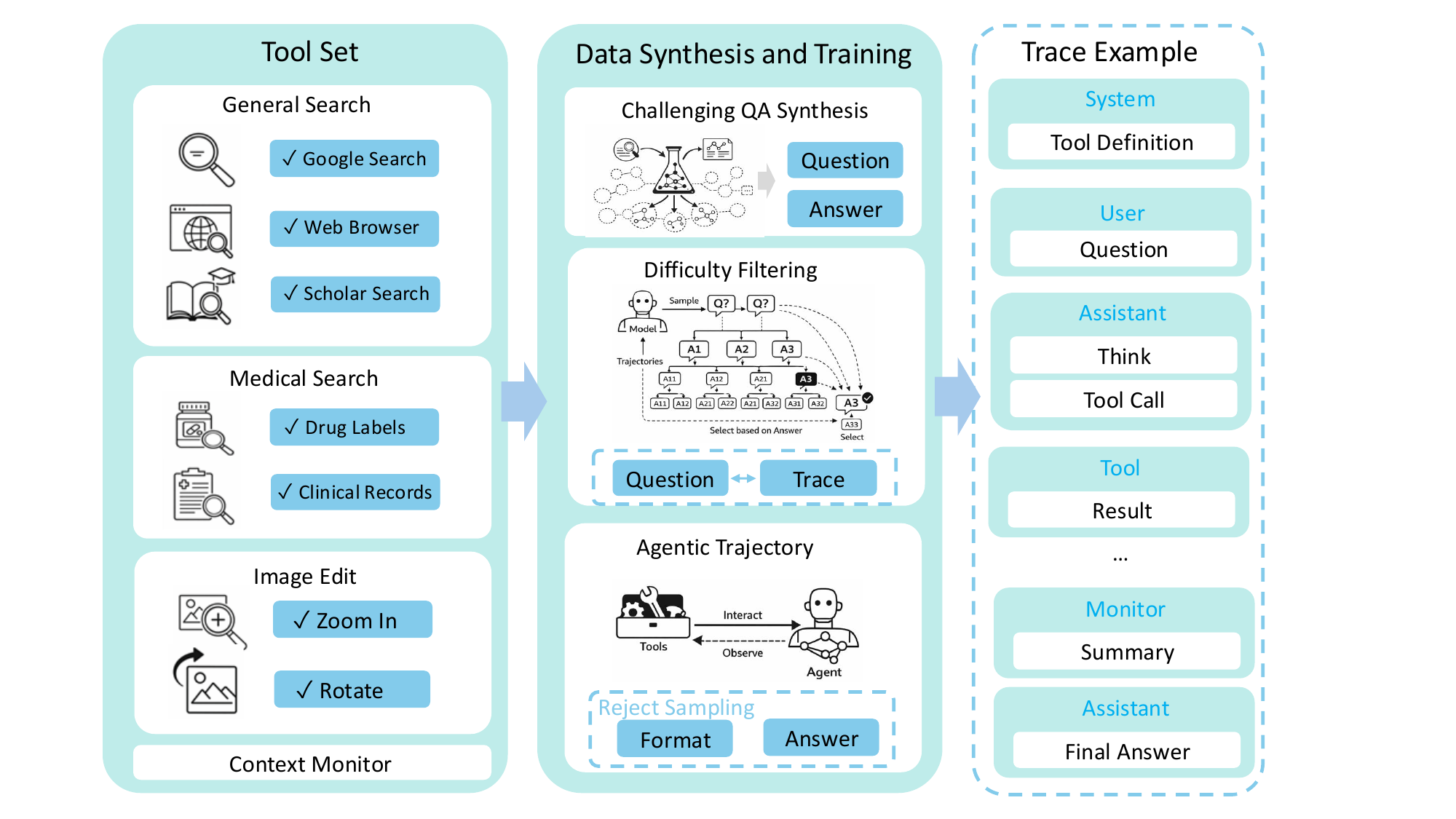}
    \caption{\textbf{Agentic Data Synthesis and Training Pipeline.} The system integrates a comprehensive toolset including General Search (Google, Scholar), Medical Search (Drug Labels, Clinical Records), and Image Editing (Zoom In, Rotate). The pipeline synthesizes challenging QA pairs through difficulty filtering, ensuring questions require multi-step tool use. Finally, an Agentic RL process trains the model to interact, observe, and summarize, producing structured traces that include specific tool definitions, thought processes, and execution results.}
    \label{fig:agentic}
\end{figure*}
\paragraph{Medical DeepResearch}
DeepResearch~\cite{team2025tongyi,lu2025deepdive} tightly couples search with reasoning: the model decomposes a problem step by step,
verifies critical details against reliable external evidence, and then produces an answer, which
helps mitigate knowledge forgetting and hallucinations.
We specialize DeepResearch for the medical domain by equipping the model with medical-specific
tools and data, enabling it to use these tools effectively in diagnostic contexts. 
We integrate standard DeepResearch tools, including \textit{General Search}, \textit{Scholar Search}, and \textit{Visit}. \textit{General Search} and \textit{Scholar Search} retrieve evidence from the web and scholarly sources given a query, while \textit{Visit} is invoked when the model decides to open a particular URL and extract the information needed, returning a concise summary of the page. 
We further introduce medical-specific tools \cite{medrax2025, tang2025medagentsbench}---\textit{Search Drug} and \textit{Search Clinical}---for drug information and clinical cases, respectively. 

General DeepResearch models are good at fact-seeking tasks but lack medical reasoning expertise. Unlike general factual QA that deals with single-hop medical entity facts, medical scenarios require multi-hop decision-making in consistent clinical contexts, including appropriate tool selection and chaining.
To tackle this, we create multi-hop medical questions and their solution paths by building multi-hop entity chains through random walks on an internal medical knowledge graph. This approach faces two core challenges:
(1) Coverage: We address it by using common diseases and drugs within each specialty as starting nodes and frequently restarting walks to avoid biased learning in small subgraphs.
(2) Relational plausibility: We solve it with an LLM-based sampling method to pick logically and clinically valid next-hop nodes from candidate neighbors.
Finally, we use an LLM to generate relevant multi-hop questions, and apply three-tier filtering to ensure their difficulty and quality. 

\begin{figure}[h]
  \centering
  \begin{tabular}{m{5cm}m{10.5cm}}
  \toprule
  \centering
  \includegraphics[width=5cm]{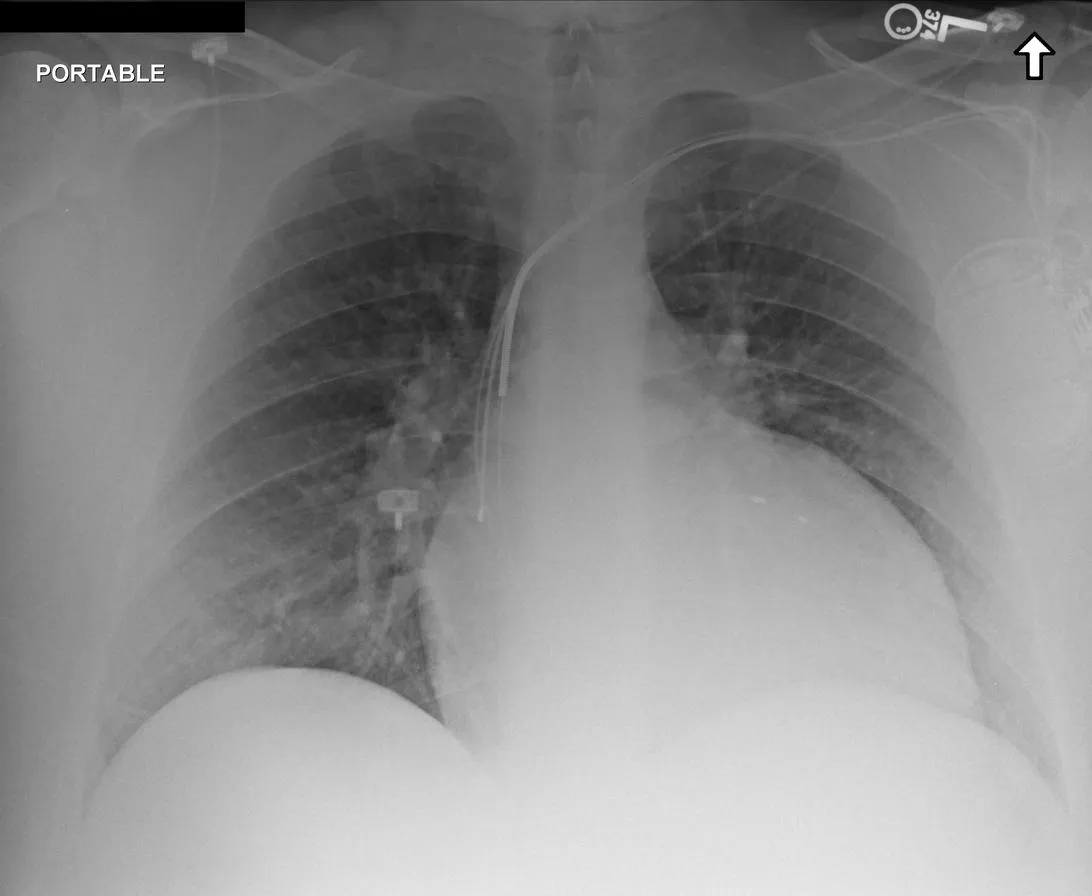}\\ 
  \vspace{1mm}
  \includegraphics[width=5cm]{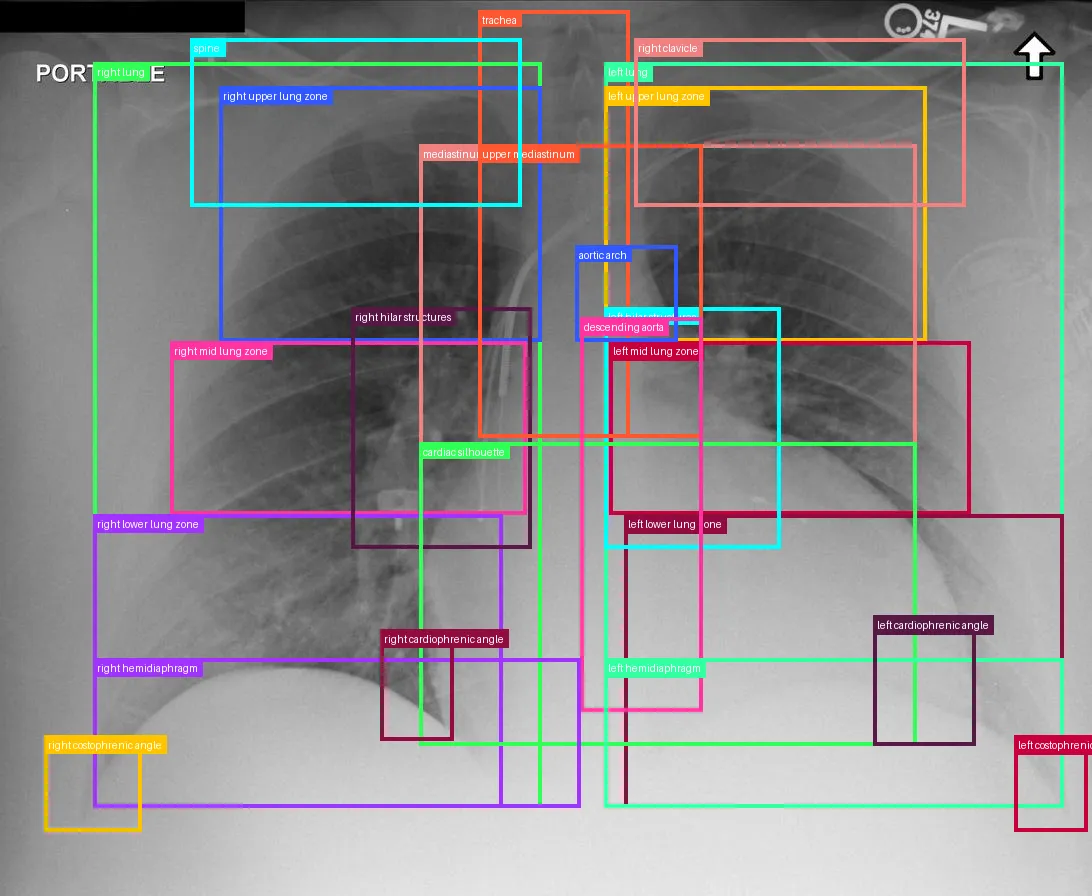}\\ 
  \vspace{1mm}
  \includegraphics[width=5cm]{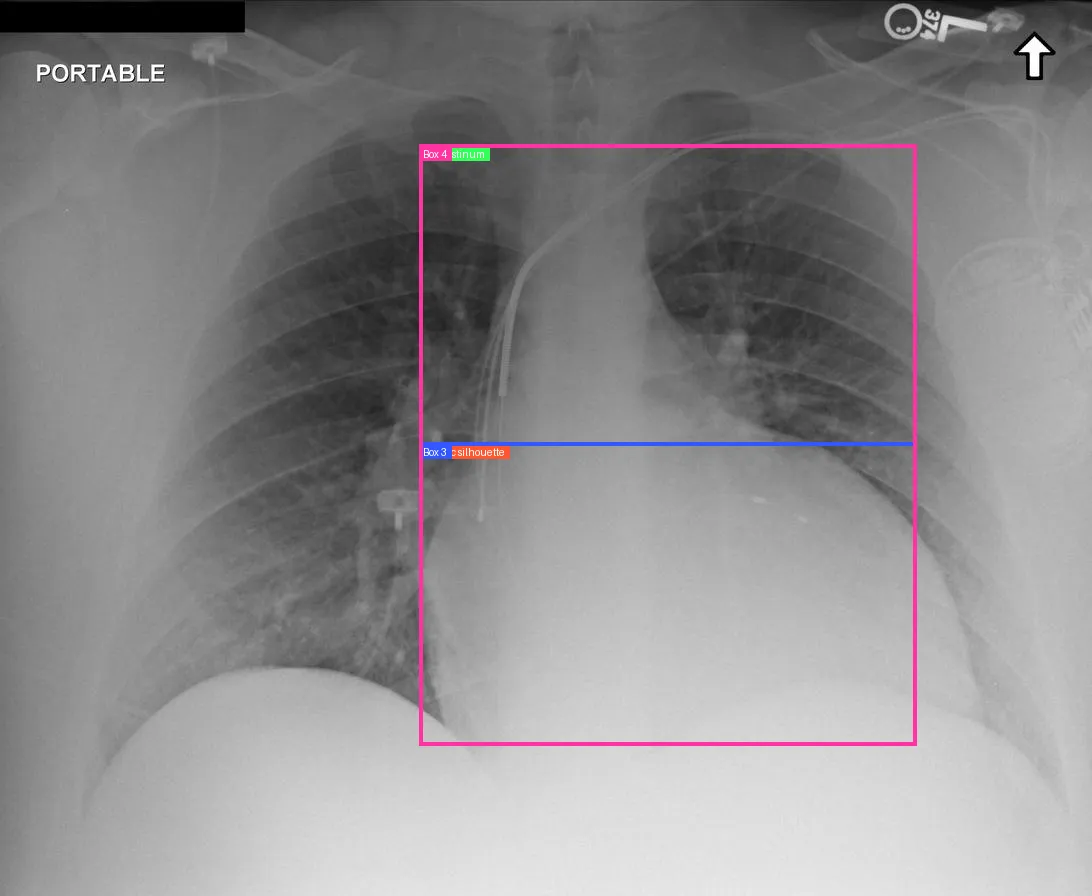}
  & 
  {\scriptsize\setlength{\parskip}{1ex}%
   \textbf{Prompts:} what abnormalities are visible in this image?\par 
   
   \textbf{Model Responses:}\par
   \texttt{<think>}\par
   Okay, I am beginning my systematic review of this chest X-ray. Starting with the airways. The trachea [478,10,629,437] appears midline with normal caliber throughout its visualized course. No deviation or narrowing is evident.
   
   Moving on to the lungs. The right lung [93,62,541,807] shows clear lung fields in all zones - the right upper lung zone [219,86,541,341], right mid lung zone [170,341,526,514], and right lower lung zone [93,514,502,807] all demonstrate appropriate aeration without focal opacities. Now looking at the left lung [604,62,1063,807] - similarly, the left upper lung zone [604,86,926,341], left mid lung zone [609,341,970,514], and left lower lung zone [624,514,1063,807] appear clear with normal vascular markings. Both right hilar structures [351,307,531,548] and left hilar structures [604,307,780,548] are within normal limits without enlargement or abnormal contour. No evidence of consolidation, nodules, or masses is seen in either lung.
   
   Now assessing the cardiovascular structures. The mediastinum [419,144,916,745] appears widened, particularly in the upper mediastinum [478,144,702,437] region. This widening could be related to vascular structures. Looking specifically at the cardiac silhouette [419,442,916,745], there's an increased transverse diameter, which suggests cardiomegaly. The aortic arch [575,245,677,341] appears prominent, which may contribute to the mediastinal widening. The descending aorta [580,317,702,711] also shows some tortuosity. These findings together suggest possible vascular congestion.
   
   Next, evaluating the diaphragms. The right hemidiaphragm [93,658,580,807] and left hemidiaphragm [604,658,1063,807] are both visible and appear to be in normal position. The right costophrenic angle [44,735,141,831] and left costophrenic angle [1014,735,1087,831] are sharp and clear without evidence of pleural effusion. The right cardiophrenic angle [380,629,453,740] and left cardiophrenic angle [873,615,975,745] also appear unremarkable.
   
   Finally, the spine [93,432,994,577] shows normal alignment. The right clavicle [190,38,521,206] and left clavicle [634,38,965,206] are intact without fractures. There are no visible foreign bodies or devices other than what appears to be a deep venous catheter coursing from the subclavian region into the superior vena cava.
   
   Conclusion: The primary abnormalities identified on this chest X-ray include cardiomegaly as evidenced by the enlarged cardiac silhouette [419,442,916,745], and mild vascular congestion affecting the mediastinum [419,144,916,745]. The lungs are clear without evidence of consolidation, effusion, or pneumothorax. The trachea is midline, and the diaphragms and costophrenic angles are normal.
   
   \texttt{</think>}\par
   \texttt{ [$\{$"finding": "Cardiomegaly", "bbox":  [419,442,916,745]$\}$, $\{$"finding": "Mild vascular congestion", "bbox":  [419,144,916,745]$\}$]}
  } \\
  \bottomrule
  \end{tabular}
\caption{Process of Medical Grounding Reasoning: The model takes the raw X-ray (top), analyzes it via logical reasoning (middle text), and outputs grounded bounding boxes for abnormalities (displayed in middle and bottom visualizations).}
\label{fig:grounding_case}
\end{figure}

\paragraph{Think with Medical Image}
To elevate the model’s visual reasoning ability in medical contexts, we introduce a specialized grounded reasoning paradigm centered on anatomical structures. 
Moving beyond generic object detection, this paradigm establishes anatomical landmarks as the fundamental anchors for vision-language alignment. It implements a structured Analyze-Reason-Conclude workflow that mirrors the clinical standard:
The analysis phase strictly adheres to the radiological protocol, systematically traversing anatomical landmarks while interleaving textual findings with precise bounding box coordinates. This mechanism explicitly binds reasoning steps to visual evidence, bridging the modality gap. Based on these grounded observations, the reasoning phase synthesizes evidence to deduce pathologies, ensuring that every diagnostic conclusion can be traced to a specific visual region. By integrating this anatomy-centered paradigm, we transform grounding from a simple detection task into a structured, evidence-based reasoning process that significantly enhances the transparency and trustworthiness of the model's clinical judgments (as shown in Fig.~\ref{fig:grounding_case}).

While anatomical reasoning provides the logical framework, visual limitations can hinder fine-grained detection. To address this, we integrated Think with Image~\cite{zheng2025deepeyes,hong2025deepeyesv2} design, enabling the model to perform secondary operations—such as rotating or zooming in—to facilitate a deeper understanding of visual information. This capability holds tremendous potential in medical imaging; for instance, zooming in on affected regions allows the model to perform granular inspection. Through exploration of tools in the multimodal domain, we identified \textit{Zoom in} and \textit{Rotate} as the most valuable primitives: the former is primarily designed for identifying subtle lesions in affected regions, while the latter assists with maintaining spatial orientation during the interpretation of complex medical images.

The training dataset is built mainly from radiology imaging data (e.g., X-rays, CT scans, MRI) and public de-identified radiology datasets, which include pathological/normal cases, metadata and expert annotations.
Given the varying diagnostic complexity and tool utility of raw radiology data, a two-stage filtering pipeline is adopted to create an effective RL training set:

\begin{itemize}
    \item Fine-grained perception filtering: A multi-agent uncertainty consensus framework is used to select cases requiring localized, region-specific perception (instead of global image context), filtering out easily solvable examples and retaining clinically challenging ones.
    \item Task-with-instrument effectiveness filtering: An iterative process assesses example difficulty with a base model (without tools) and checks if tool-augmented inference improves correctness, retaining only tool-effective cases. This step is interleaved with training to adapt to the model’s evolving competence.
\end{itemize}

This filtering pipeline produces a high-quality RL dataset focused on diagnostically challenging, tool-enhanced cases, guiding the model to learn appropriate tool use for effective clinical reasoning. 

\subsection{Training Recipe}

To effectively synthesize the heterogeneous cognitive architectures into a resilient clinical system, we implemented a unified, multi-objective curriculum learning framework. This consolidated strategy is explicitly targeted at simultaneously resolving the deficit in fundamental reasoning capabilities and the destructive interference between perceptual fidelity and abstract reasoning. 
Our approach orchestrates a systematic evolution that fuses foundational logic, adaptive execution, and perceptual grounding within a single comprehensive training phase through the following strategic pillars:

\begin{itemize}

    \item \textbf{Curriculum Learning.} 
    We implement a progressive training strategy to build medical reasoning capabilities, advancing from foundational logic initialization to complex long-horizon deduction, then to perceptual-logic alignment, and finally to adaptive hybrid fusion. This hierarchical approach mimics human cognition to systematically resolve core limitations: missing reasoning baselines, logical discontinuities in complex tasks, and the trade-off between perception and reasoning.

    \item \textbf{ViT Joint Training Optimization.} 
    Medical VQA relies heavily on distinguishing minute visual discrepancies that frozen encoders might overlook. We unfreeze the visual backbone to enable joint optimization with the language model. This ensures that the extracted visual features are not static, but instead adapt to detect density changes and subtle anomalies specific to medical images.

    \item \textbf{Progressive Distribution Warm-up.}
    To mitigate the risk of catastrophic forgetting and capability imbalance caused by drastic distributional shifts between pre-training and mid-training, we implemented a progressive adjustment strategy. Rather than an abrupt regimen change, we gradually anneal the proportion of target reasoning data throughout the training steps. By incrementally increasing the density of complex medical reasoning samples while maintaining a replay buffer of general-domain data, we effectively smoothed the distribution fluctuations, allowing the model to securely anchor new capabilities without eroding its original competencies.

    \item \textbf{Balancing Perception and Reasoning.}
    To reconcile the verbose logic of complex derivations with fine-grained perception, we implemented a dual-track alignment strategy featuring anatomy-centered grounding.
    Furthermore, to transcend the limitations of closed-system parameterization, we incorporated tool-augmented trajectories derived from medical knowledge graphs, transforming the model into a reasoner that maintains perceptual fidelity while verifying critical details with external evidence.

\end{itemize}

\section{Post-training}
\label{sec:posttrain}

\begin{figure*}[!h]
    \centering
    \includegraphics[width=0.9\linewidth]{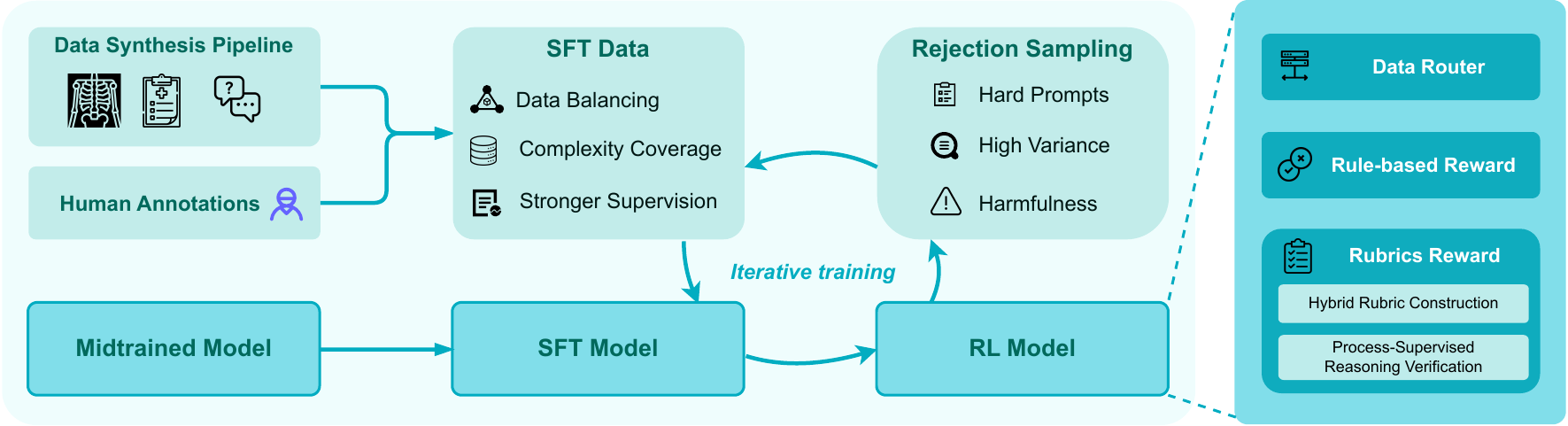}
    \caption{\textbf{Post-training Pipeline}. During post-training, we adopt the same synthesis pipeline as in mid-training, augment it with expert annotations, and perform SFT followed by RL optimization. In parallel, we mine hard negatives via rejection sampling and feed them back into the training loop for iterative refinement.}
    \label{fig:post_train_pipeline}
\end{figure*}

The core objective of post-training is to establish foundational clinical capabilities, focusing on accuracy and regulatory compliance through SFT. We also aim to improve reasoning in complex medical scenarios through RL, ultimately achieving reliability suitable for clinical deployment. The overall post-training pipeline is illustrated in Fig.~\ref{fig:post_train_pipeline}.


\subsection{Supervised Fine-Tuning}\label{sec:con_of_sft}
Supervised fine-tuning (SFT) uses the highest-quality data in the training pipeline and typically provides broad task coverage and strong instruction following capabilities, distinguishing it from mid-training.

\subsubsection{SFT Data Construction}

\paragraph{Human annotations}  
Our human-preference data are triggered primarily by a multi-expert consistency mechanism. For the same image–text prompt, we query multiple mutually independent VLMs to produce candidate responses, and then aggregate and score these candidates using an expert-consistency metric. If model disagreement is substantial, or if key medical entities and conclusions cannot be reliably determined from consistency signals, the sample is escalated to a human annotation workflow. During annotation, medically trained annotators consult the multi-model outputs and their consensus evidence to fact-check candidate answers and provide a preference ranking, yielding high-quality preference pairs.

\paragraph{Synthetic data}  
Our synthetic preference data follows the closed-loop synthesis pipeline used in the mid-training stage (Fig.~\ref{fig:mid_training_overview}), with additional enhancements for diversity and robustness. Specifically, we perform prompt rewriting and generate expression variants without altering medical facts or ground-truth answers. Meanwhile, a data-balancing module conducts coverage-based sampling over the ``site/disease/lesion $\times$ difficulty'' grid, and applies targeted oversampling to necessary long-tail and high-risk regions. As a result, the synthetic preference signal achieves both balanced coverage of overall capabilities and stronger supervision on error-prone critical points.

\subsubsection{Instruction Following}

Instruction following reflects whether a model can correctly interpret and execute user requests under constraints such as task scope, formatting, multi-turn context, and safety rules. In clinical settings, this requirement is stricter because instructions are often long-context, multi-turn, and may include implicit constraints or conflicts. Therefore, our SFT stage explicitly targets robust medical instruction following, enabling users to switch task paradigms via configurable prompts.
A prerequisite for reliable instruction following is accurate instruction understanding. The model must correctly parse explicit rules and infer implicit requirements, especially under layered constraints. We improve this capability from both the data and method perspectives.

\paragraph{Instruction Data Analysis and Synthesis}
We collect diverse medical tasks from open-source datasets and online resources, and then use LLMs to expand them in depth and breadth. These tasks are standardized into an atomic instruction set. We further sample conflict-free combinations of atomic instructions to construct both single-turn and multi-turn instruction following data, ensuring coverage of departments, task types, and constraint patterns.
Real-world medical requests often contain hidden reasoning steps rather than explicit instructions. To model this, we build a graph-structured representation that links knowledge nodes, relations, and operations. We then convert the graph into natural-language instructions that preserve implicit reasoning requirements, and use the graph as a reference to generate structured rationales for training.

\paragraph{Instruction Understanding and Reasoning} 
Existing models are prone to non-compliance under strict medical constraints, which may lead to safety risks (e.g., violating prohibitive rules). We first conduct large-scale SFT with the synthesized instruction data to strengthen basic compliance. However, forward-only synthesis can introduce subtle instruction–response misalignment and makes quality control difficult. We therefore adopt a reverse construction strategy: start from high-quality clinician-like responses, infer the most compatible instructions and form high-quality instruction–response pairs, improving instruction parsing and constraint adherence.

To further strengthen reasoning under constraints, we introduce structured ``thinking'' supervision in SFT. We generate two types of rationales: (1) prompt-guided structured rationales with explicit constraint checking, and (2) rationales expanded from the explicit reasoning graph to encourage fine-grained logical decomposition.

\subsection{Reinforcement Learning}

Following data engineering, mid-training, and SFT, the model possesses extensive domain knowledge but lacks the proficiency to fully mobilize it for complex clinical reasoning. To bridge the gap between latent competence and practical performance, we implement a comprehensive Reinforcement Learning (RL) stage. 

\subsubsection{RL Data Construction}\label{subsubsec:data_synthesis_pipeline}
We follow the data production logic established in the SFT stage and additionally introduce a data-filtering module to select high-value samples from the candidate pool, e.g., complex cases with SFT accuracy in the 60\%–80\% range, so as to prioritize coverage of the model's difficult regions. For samples that require LLM-as-a-Judge evaluation, we further extract and construct high-quality, discriminative atomic evaluation points to ensure that the reward signal is clear and reliable. Building on this, we augment the corpus with RL-specific incremental data, including tool-assisted reasoning instances and conflict scenarios in multi-turn dialogues, to strengthen robustness and instruction consistency in complex settings.

\subsubsection{Multi-Layered Hybrid Reward System}\label{subsubsec:multi_layered_hybrid_reward_system}

\begin{figure}[htbp]
    \centering
    \includegraphics[width=\linewidth]{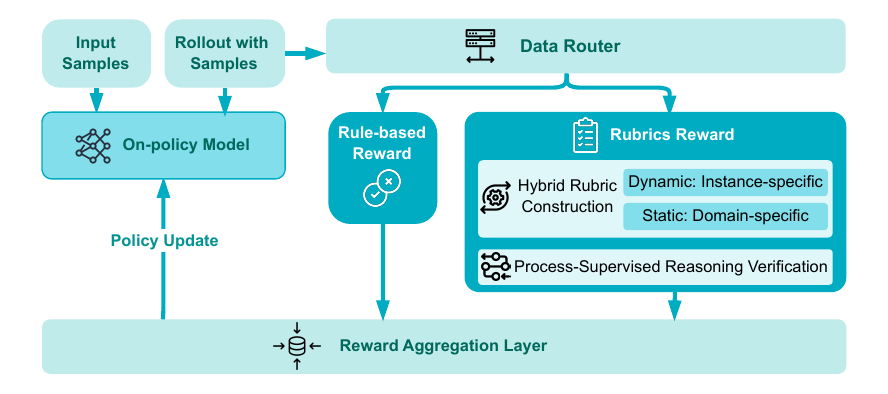}
    \caption{\textbf{Architecture of the Multi-Layered Hybrid Reward System.}
    Data Router directs inputs to two parallel reward modules: Rule-based Reward, and Rubric Reward. The Aggregation Layer combines these signals into a final scalar reward for RL optimization.}
    \label{fig:posttrain_rl_reward_system}
\end{figure}

The Multi-Layered Hybrid Reward System implements a hierarchical architecture that systematically validates, routes, and scores model outputs through specialized evaluation modules (Fig.~\ref{fig:posttrain_rl_reward_system}). This framework begins with distributing samples to task-appropriate graders, ultimately synthesizing diverse reward signals into a unified optimization objective that balances clinical accuracy, logical coherence, and safety alignment.

\textbf{Data Router.} The Data Router dynamically directs input samples to specific evaluation pipelines based on metadata. Instead of a monolithic approach, it interprets prompt intent to select the appropriate Grader—routing alignment tasks to the Rule-based and the Rubrics reward model. This ensures reward signals are contextually tailored to specific training objectives.

\textbf{Rule-based Reward.}
This deterministic evaluator handles tasks with well-defined ground truth that can be verified through exact string matching or regular expression patterns. Employing set-theoretic principles, it compares model predictions against reference answers, awarding proportional credit for partial overlaps and maximum scores for exact matches. All calculations are normalized to a unified scale to ensure compatibility within the broader optimization framework.

\textbf{Rubrics Reward.} The Rubric Reward mechanism provides a unified verification framework that evaluates candidate responses against multidimensional assessment criteria, ensuring comprehensive alignment with both contextual requirements and substantive quality standards. To establish reliable scoring anchors, we generate high-quality reference responses through best-of-N sampling from expert models fine-tuned on domain-specific datasets curated and labeled by human specialists.

\textit{Hybrid Rubric Construction.} We establish evaluation rubrics through a dual-pathway approach combining dynamic context adaptation with consensus-based quality standards. The dynamic pathway generates instance-specific rubrics: an Expert Analyst model analyzes the system prompt, dialogue history, and user query to construct tailored evaluation criteria capturing contextual requirements such as logical coherence, empathetic expression, and intent alignment. The static pathway establishes domain-specific gold standard rubrics through multi-model consensus filtering followed by human expert refinement, encompassing critical assessment dimensions—for instance, Medical Knowledge and Patient Experience in medical dialogue systems. For tasks with explicit reference answers, rubrics incorporate semantic alignment criteria where LLMs/VLMs assess meaning equivalence beyond surface-level lexical matching. Candidate responses are scored against both dynamic contextual rubrics and static domain rubrics, enabling the model to satisfy immediate conversational demands while maintaining substantive quality benchmarks.

\textit{Process-Supervised Reasoning Verification.} To enhance multi-step reasoning capabilities, we introduce dense supervision of the model's internal deliberation process encoded in \texttt{<think>} blocks. A dedicated evaluation module assesses this intermediate chain of thought across three dimensions: framework completeness (whether all relevant constraints are identified), logical robustness (validity of inference steps), and exploratory depth (thoroughness of solution space examination). This process supervision compels the model to explicitly analyze task constraints and formulate execution strategies before generating responses, substantially improving performance on complex reasoning tasks.

To mitigate the instability inherent in pointwise scoring for nuanced medical criteria, we adopted Reference-Augmented Rubric Evaluation. By anchoring assessments to gold-standard references, this framework reformulates evaluation as a pairwise semantic matching task. This methodology significantly reduces variance, ensuring that clinical reasoning is validated against concrete ground truth rather than abstract definitions.
Building on this robust baseline, we apply Soft Shaping to optimize auxiliary quality metrics using continuous multipliers that scale the base evaluation scores based on non-core criteria like response length, formatting, and Pharmacist Evaluation. These multiplicative adjustments compound penalties for deviations while preserving value for high-quality outputs, creating smooth optimization gradients that drive balanced performance without compromising semantic accuracy.

\paragraph{Reward Signal Fusion}
We synthesize a unified optimization objective by non-linearly fusing criterion-based medical scoring, fundamental evaluations, and reasoning process rewards with strict safety constraints. This formula is governed by a binary gating signal that nullifies the entire reward if critical safety rules are violated. The core reward aggregates these components—along with a partial order correction against expert baselines—using dynamically computed weights to balance task characteristics. Specifically, the dominant rubrics-based score is further modulated by soft shaping multipliers, ensuring the final signal rigorously promotes clinical accuracy, logical coherence, and safety alignment.

\subsubsection{RFT-Enhanced Curriculum Reinforcement Learning}\label{subsubsec:iterative_curriculum}

Training medical agents presents a fundamental challenge: reconciling heterogeneous multi-modal data—spanning basic instruction following to long-horizon clinical reasoning over high-resolution imaging—under layered, expert-driven, safety-constrained rewards (§\ref{subsubsec:data_synthesis_pipeline}, §\ref{subsubsec:multi_layered_hybrid_reward_system}). This heterogeneity creates conflicting optimization pressures that general method cannot resolve. Our preliminary experiments reveal that simultaneous training (Fusion Paradigm) induces gradient conflicts and capability oscillations, while sequential training (Staged Paradigm) causes entropy collapse, where overconfidence on simple tasks eliminates plasticity for complex reasoning. We therefore propose the RFT-enhanced Iterative Curriculum strategy to address these limitations.

\textbf{Iterative Curriculum Strategy.}
To address the limitations of both the fusion and staged paradigms, we implement an Iterative Curriculum strategy grounded in Pareto multi-task learning theory that alternates between supervised capability distillation and curriculum-based policy optimization. This approach frames the training process as a multi-objective Pareto optimization system operating under dynamic constraints. Rather than attempting to resolve data heterogeneity through static mixing or rigid sequencing, we leverage it as a controllable variable within a dynamic curriculum where rejection sampling fine-tuning (RFT) and reinforcement learning create synergistic improvement cycles.

\textit{Iteration.} Each iteration comprises a four-phase progression that implements two core principles: mitigating gradient conflicts through focused micro-cycles that prevent simultaneous destabilization of parameter updates, and regulating entropy via periodic re-injection of foundational data to preserve model plasticity across the capability spectrum. 
The RFT Phase initiates each cycle by converting the model's latent competence into systematic supervision: we sample multiple responses from the current policy for each training instance, identify cases where at least one output meets our multi-layered reward criteria, and distill these superior responses back into the training corpus, systematically converting sporadic successes into reliable behavioral cloning signals. 
Following this capability distillation, the model proceeds through three complementary RL curriculum phases that form a coherent learning cycle: 
the Foundation Phase establishes reward stability through short-context instructions and simple multimodal samples, anchoring gradients while maintaining core instruction following capabilities; 
the Specialization Phase then targets high-difficulty clinical reasoning by substantially upweighting complex samples—long-horizon cases and high-resolution imaging to push performance boundaries on capabilities that failed to emerge under staged training due to premature entropy collapse; 
finally, the Alignment Phase reintroduces general domain and safety-constrained samples as both regularization and plasticity preservation, preventing catastrophic forgetting while maintaining safe distributional boundaries and addressing the overconfidence observed in preliminary staged experiments.

\textit{Benefit.} This RFT-driven capability distillation creates synergistic support across the three-phase RL curriculum. During the Foundation Phase, RFT-enhanced data enables stable gradient anchoring by providing cleaner behavioral cloning signals, allowing the model to efficiently internalize correct reasoning patterns rather than learning from noisy generation distributions. In the Specialization Phase, improved supervision facilitates effective reward bootstrapping on complex clinical cases: the model can reliably generate intermediate-quality responses that receive positive reinforcement, enabling productive exploration of long-horizon reasoning and high-resolution imaging analysis that would otherwise yield sparse rewards. During the Alignment Phase, the reduced distribution shift between RFT-supervised and RL-optimized policies prevents catastrophic forgetting by ensuring regularization operates on a coherent capability manifold, while also preserving policy diversity to prepare the model for subsequent RFT iterations as frontier capabilities expand.

 Through iterative cycling, the model achieves stepwise improvement on the hybrid reward signal, progressively integrating capabilities without the gradient interference of simultaneous training or the inflexibility of sequential paradigms. However, two fundamental challenges persist due to medical reasoning data characteristics: the cold start problem in early iterations, where sparse rewards from complex clinical conditions hinder bootstrapping of valid reasoning chains, and exploration stagnation in later iterations, where accumulated confidence suppresses discovery of alternative solution paths. We address these through two stage-specific algorithmic interventions:

\textbf{Dynamic Hint-based Curriculum} addresses the cold start problem by providing structured scaffolding during early iterations. We inject hierarchical hints—ranging from visual perception cues to intermediate logical steps—into the input space, serving as gradient primers that transform sparse reward signals into dense learning opportunities. These hints enable the model to bootstrap valid reasoning chains even when facing novel complex tasks. Across subsequent iterations, we apply a linear decay schedule to hint density, forcing a gradual transition from external assistance to internalized parameters, ensuring zero-shot capability by intermediate stages.

\textbf{Entropy-aware Adaptive Regulation} counters exploration stagnation by dynamically modulating the ``exploration versus exploitation" trade-off in later iterations. We integrate a real-time monitoring system that tracks policy entropy as a proxy for reasoning diversity. When low entropy periods are detected—indicating overconfidence in familiar patterns—the system dynamically introduces an entropy bonus into the loss function. This mechanism prevents premature convergence to local optima, reinvigorating exploration of alternative reasoning paths and ensuring continuous optimization even as performance approaches the ceiling.

\section{Unified Med-VLM Benchmark}
\label{sec:unified_evaluation}

Medical vision-language models are increasingly positioned as foundation components for clinical-facing systems, spanning tasks from fine-grained visual inspection\cite{yumedlesionvqa, gong2025med, mmmu2024, mmmupro2025} and radiology understanding~\cite{vqarad2018, slake2021, omnimedvqa2024} to medical dialogue~\cite{Zhu2025DiagnosisArena, Zuo2025MedXpertQA, healthbench2025}, documentation and report writing~\cite{mimiccxr2019, chexpertplus2024, iuxray2016}. 
However, evaluation practice in the medical VLM literature remains fragmented: different works adopt different benchmark subsets, prompting styles, scoring scripts, and data hygiene assumptions, which makes cross-paper comparisons brittle and slows down community convergence on what ``reliably better'' means in medicine. 
This is visible across recent representative systems \citep{sellergren2025medgemma,medgemini2024,quarkmed2024,lingshu2024}, where reported numbers are often not directly comparable due to protocol-level mismatches rather than model differences. 

To address this, we introduce a Unified Med-VLM Benchmark. The suite consolidates 30+ public benchmarks and pairs them with a standardized evaluation harness.

\paragraph{Design goals}
We build the suite around three practical goals. First, capability coverage: medical models can fail in many different ways, so the benchmark suite should reflect a broad capability surface rather than a single leaderboard. Second, protocol reproducibility: evaluation results for the same model checkpoint should be stable across runs and robust to minor formatting variations. Third, deployment relevance: beyond answer correctness, we emphasize behaviors that determine usability in real-word products, such as multi-turn consistency, safety-aware responses, and long-form report generation.

Across benchmarks, we normalize evaluation with a consistent harness:
\begin{itemize}
  \item \textbf{Task normalization:} each dataset is mapped into a small set of task families (MCQ, short QA, long QA/dialogue, report/caption generation, OCR, agentic search/decision).
  \item \textbf{Prompt templates:} we use capability-aware templates that standardize instruction phrasing, answer format constraints, and the placement of auxiliary context.
  \item \textbf{Answer parsing:} for closed-form tasks (MCQ/short QA), we enforce deterministic parsing (option extraction, numeric normalization, whitespace/punctuation normalization). For open-form tasks (dialogue/report/caption), we score with task-appropriate metrics and/or rubric-style structured judging to distinguish clinical correctness, completeness, and unsupported claims.
  \item \textbf{Reporting:} we report per-benchmark scores and also provide category-level summaries (macro-averaged across benchmarks within a category) to avoid a single large dataset dominating the overall picture.
\end{itemize}
This protocol turns a diverse set of medical evaluations into a coherent measurement system, while still respecting each benchmark's original intent.

\subsection{Public Benchmarks}
\label{subsec:public_benchmarks}

We organize public benchmarks into six capability categories. Each category has its own table, and each table is referenced in the corresponding capability description to make the mapping explicit and auditable.

\paragraph{Visual Diagnosis \& Image Recognition}
This category targets clinical visual understanding broadly, including recognizing visually salient disease cues and answering multimodal questions that require grounding in the image rather than text-only recall. 
We include medical subsets of comprehensive multimodal benchmarks to stress generalization across diverse medical visual concepts, as in Tab.~\ref{tab:public_visual_diagnosis}.

\paragraph{Medical Imaging}
Medical imaging evaluation emphasizes radiology and pathology perception, multi-scale cues (organ-level, lesion-level, pattern-level), and multimodal fusion between imaging evidence and clinical knowledge. We include benchmarks spanning multiple modalities and task types to measure both breadth (coverage) and depth (reasoning under domain constraints) (Tab.~\ref{tab:public_medical_imaging}).

\paragraph{Diagnosis}
Diagnosis benchmarks focus on end-to-end clinical reasoning: synthesizing evidence, performing differential diagnosis, handling uncertainty, and generalizing to rare conditions. Compared to exam-style QA, many diagnosis tasks are closer to real clinical narratives and therefore stress robustness to distractors and the ability to justify conclusions from evidence. Tab.~\ref{tab:public_diagnosis} summarizes the diagnosis benchmarks we include.

\paragraph{Medical Text}
Medical text benchmarks measure professional knowledge coverage and text-only clinical reasoning, including licensing-exam formats, biomedical QA, and multi-turn medical dialogue. This category captures common real-world failure modes such as ``missing knowledge with overconfident reasoning'' on dosage, contraindications, and guideline questions. We include both Chinese and English evaluations to measure multilingual robustness (Tab.~\ref{tab:public_medical_text}).

\paragraph{Medical Report}
Medical report generation stresses long-form coherence and, more importantly, faithfulness: the ability to describe findings without introducing subtle hallucinations or unsupported claims. Evaluation in this category is particularly sensitive to protocol choices (e.g., what constitutes an omission versus an incorrect addition), which motivates our unified prompting, structured output requirements where applicable, and complementary inhouse faithfulness tests. Tab.~\ref{tab:public_medical_report} lists the report benchmarks included.

\paragraph{Instruction Following}
Instruction-following evaluation measures controllability in long medical dialogues: constraint satisfaction, memory, consistency across turns, and handling of implicit requirements (e.g., multi-step instructions embedded in patient-facing interactions). These behaviors often determine whether strong medical knowledge translates into a stable user experience under real system prompts. Tab.~\ref{tab:public_instruction_following} summarizes the benchmarks used.

\subsection{Inhouse Benchmarks}
\label{subsec:inhouse_benchmarks}

Public benchmarks remain indispensable but do not fully cover several deployment-critical axes, particularly for real-world patient images and medical documents. 
We therefore introduce targeted inhouse evaluations that complement the public suite while following the same unified protocol. 
These benchmarks are designed to be high-signal ``gap detectors'': they emphasize capabilities that frequently cause failure in practical systems yet are under-measured in the public landscape.

\paragraph{Inhouse VQA}
We include an inhouse VQA benchmark focused on real-world clinical images where lesion cues and disease regions must be recognized reliably. The benchmark contains more than 100k questions and fully aligns with real-world clinical applications.
This benchmark complements the broader public visual-diagnosis probes by testing deployment-like imagery and question styles.

\paragraph{Inhouse OCR}
Inhouse OCR is constructed from real, human-captured images of medical reports, deliberately retaining perspective distortion, blur, shadows, glare, partial occlusion, and cluttered backgrounds to reflect patient and clinician workflows in routine practice. The benchmark evaluates not only character-level recognition accuracy but also the real-world ability of medical AI assistants to recognize and interpret reports.

\paragraph{Inhouse Caption}
We introduce an inhouse caption benchmark built from real-world clinical images focusing on lesions and other visually grounded findings, drawn from specialties where image evidence is essential for diagnosis (e.g., dermatology and dentistry). For each image, we curate a structured set of clinically meaningful “key points,” including lesion-focused description, likely diagnosis, and management recommendations. To ensure reliability, the correctness of all key points is cross-validated by two licensed physicians through independent review and adjudication. Building on these verified key points, we propose an automatic scoring scheme for evaluating model-generated captions: correctly covered key points receive positive credit, while incorrect statements are penalized. This reward-and-penalty design encourages clinically faithful, high-utility captions and enables fine-grained comparison of models under a consistent, medically grounded rubric.

\section{Conclusion}
\label{sec:conclusion}

In this paper, we presented \doubao{}, a medical vision-language foundation model designed for strong multimodal understanding and generation in real-world healthcare scenarios. Our results show that \doubao{} is competitive across a broad range of medical benchmarks and exhibits robust integrated capabilities across medical text, medical images, and clinical understanding.  

Beyond benchmark performance, \doubao{} demonstrates generalization to practical medical workflows that require grounding outputs in evidence and handling multi-step reasoning with external tools.  These behaviors are important for moving from isolated tasks toward systems that can support clinicians in realistic settings.

Looking forward, our analysis suggests several promising directions. First, scaling training data and compute—together with stronger evaluation—remains an immediate path to improving performance and robustness.  Second, we identify limitations shared by current medical VLMs, including mitigating hallucinations in long-form generation, improving reliability under distribution shift, and strengthening multi-step medical reasoning.  Addressing these challenges is a core part of our ongoing work, including deeper evidence-grounding, more reliable tool-use, and broader coverage of medical concepts and modalities. 

Overall, we hope \doubao{} and the accompanying evaluation framework will help advance the development of trustworthy medical multimodal foundation models and encourage more standardized, clinically meaningful assessment going forward.

\clearpage

\bibliographystyle{plainnat}
\bibliography{main}

\clearpage
\section{Contributions}
\label{sec:contributions}
The authors are listed in alphabetical order by their first names.
\subsubsection*{Contributors}
\begin{multicols}{4}
\raggedcolumns
\small

Bahabaike Jiangtulu\\
Baorong Shi\\
Bo Cui\\
Boyuan Jiang\\
Changkun Liu\\
Chao Gao\\
Deli Yu\\
Fang Qian\\
Haihua Yang\\
Hao Wang\\
Huichao Wang\\
Jiale Chen\\
Jianfei Pan\\
Jianghang Lin\\
Jiaying Zhao\\
Jieqiong Cao\\
Jinghao Lin\\
Kai Ruan\\
Kai Wu\\
Qian Xu\\
Lin Yang\\
Shengsheng Yao\\
Shengzhi Wang\\
Tao Chen\\
Xiaojun Xiao\\
Xiaozhong Ji\\
Xu Wang\\
Yifan Zhang \\
Yijun He\\
Yiwen Ye\\
Yonggang Yang\\
Yuancheng Yang\\
Yuhang Wu\\
Zhaolong Huang\\
Zhixiong Yang\\
Zhongkai Zhao\\
Zihan Wang\\
Ziqi Zhou\\

\end{multicols}

\clearpage

\beginappendix

\section{Qualitative examples}
\label{sec:app:qualitative}

\subsection{Medical DeepResearch}
\label{sec:app:rebus}

\textcolor[HTML]{2E5AA8}{\paragraph{Milestones for Medical Deepresearch}}
Our team has achieved \textbf{\#1 place} in both the
Internal Reasoning and Agentic Reasoning tracks of \textbf{Cure-Bench@NeurIPS 2025} (\url{https://curebench.ai/}), which is a competition and benchmark for evaluating AI models in drug decision-making and treatment planning hosted by Harvard Medical School in NeurIPS 2025.

\begin{figure}[!ht]
  \centering
  \begin{tabular}{c}
    \toprule
    \includegraphics[width=0.95\textwidth]{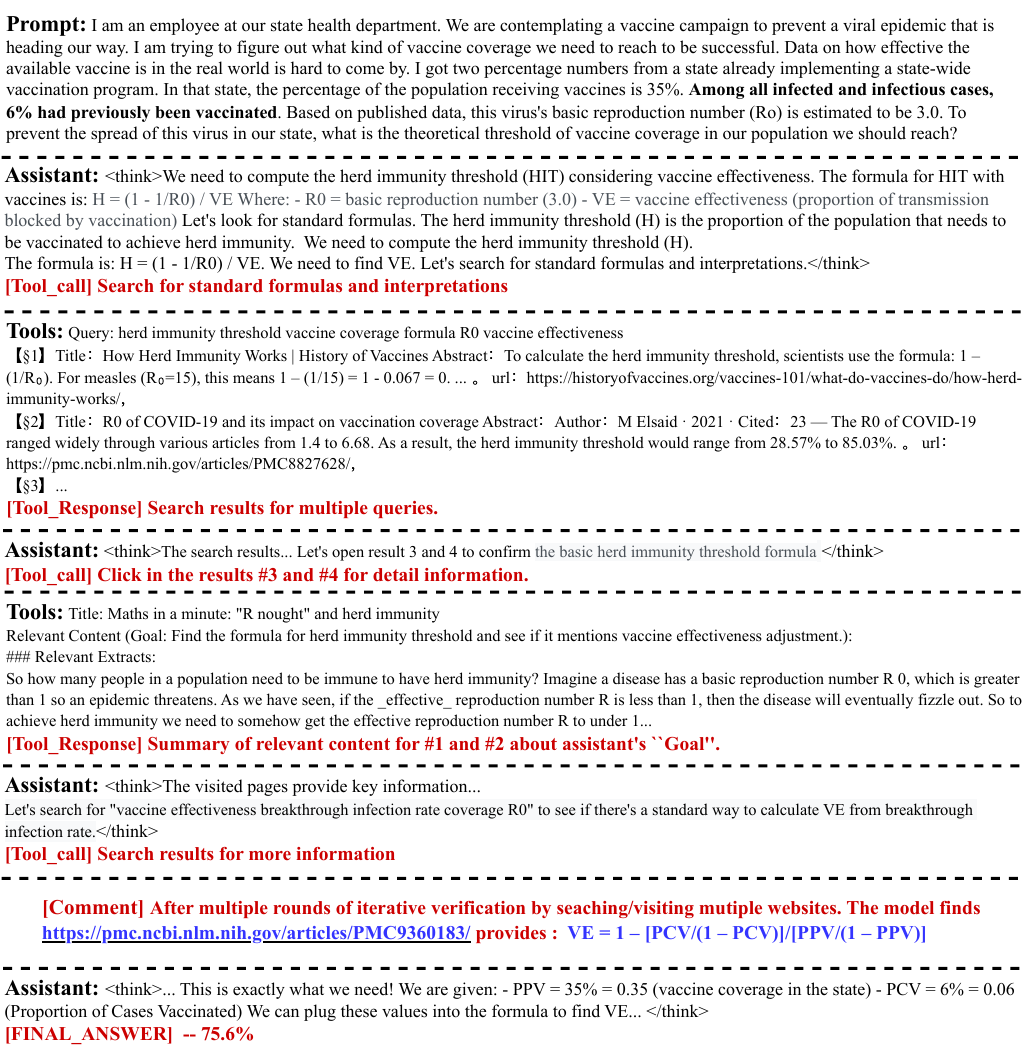}\\
    \bottomrule
  \end{tabular}
\caption{An example of tool-assisted medical reasoning: the model retrieves and verifies evidence before producing the answer.}
\label{fig:tool-assisted-reasoning}

\end{figure}

\clearpage 

\subsection{Think with Medical Image}
\label{sec:thin_with_image_sft}
\begin{figure}[!ht]
  \centering
  \small 
  \renewcommand{\arraystretch}{1.2} 
  
  \begin{tabular}{p{0.96\linewidth}}
    \toprule
    
    \textbf{Prompt:} Which diseases are visible in the image? \\
    \vspace{2mm}
    {\centering \includegraphics[height=3.5cm]{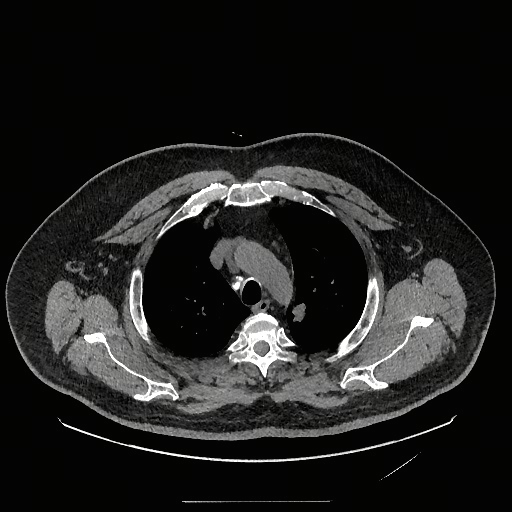} \par}
    
    \vspace{2mm}
    \vspace{2mm}
    
    \textbf{Model Responses:}\par
    
    \texttt{<think>} 
    The user now needs to follow the analysis steps and first use the ZOOM tool to enlarge the regions in the chest CT that may contain disease. There appear to be two suspected lesions in the image, so both suspicious areas need to be magnified for closer inspection.
    First, determine that ``imgidx" is 0 (since this is the first image in the current session). The ``bbox\_str" should be chosen to include the mediastinum or other potentially abnormal regions.
    The first suspicious area is located in the mediastinal region. Set ``imgidx" to 0, ``label" to ``lesion1\_calcification", and temporarily set ``bbox\_str" to ``384 414 585 583". The cropping range should be slightly expanded to avoid missing the boundary. Set ``scale" to 2 to facilitate evaluation of whether this represents calcification.
    The second suspicious area is located in the peripheral lung region.
 
    ...
    
    allowing clearer assessment of its margin characteristics and distribution features. 
    \texttt{</think>}\par
    
    \vspace{2mm}
    
    \noindent\texttt{<tool\_call>} 
    ``arguments": `` \{``label": ``lesion1\_calcification", ``imgidx":0, ``bbox\_str": ``<bbox>384 414 585 583</bbox>", ``scale":2\}", ``name": ``ZOOM", ...
    \texttt{</tool\_call>}\par
    
    \vspace{2mm}
    

    \begin{center}
    \includegraphics[height=3.3cm, keepaspectratio]{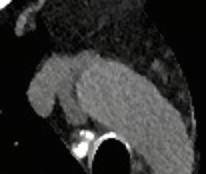}
    \hspace{0.5em}
    \includegraphics[height=3.3cm, keepaspectratio]{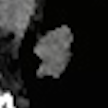}
    \end{center}
    
    \vspace{2mm}

    \texttt{<think>} 
    After enlarging the two regions separately using the ZOOM tool, the first mediastinal lesion appears as a high-density focus with relatively clear boundaries. Its shape and density are more consistent with the characteristics of a calcified lesion.
    The second lesion is located in the peripheral lung. It presents a lobulated shape with irregular margins (possibly with spiculations), and its distribution is not along the bronchial pathways, which are imaging features suggestive of peripheral lung cancer.
    ...
    \texttt{</think>}
    After magnified observation of the two suspicious areas, the first is identified as a mediastinal calcification, while the second appears as a lobulated mass in the peripheral lung with irregular edges and no bronchial distribution, highly suspicious for peripheral lung cancer.
    In summary, the diseases shown in the image are: mediastinal calcification + peripheral lung cancer. 
    \par
    
    \vspace{2mm}
    
    \textbf{Answer}: Lung Cancer \\
    \bottomrule
  \end{tabular}
  
  \caption{An example of magnification-assisted reasoning. The process flows vertically: the model analyzes the original image, calls a zoom tool, observes the zoomed-in crop (inserted in the middle), and concludes the diagnosis.}  \label{fig:zoom-in-operation}
\end{figure}

\clearpage

\subsection{Think with Grounding}
\label{sec:Grounding Reasoning}

\renewcommand{\arraystretch}{1.2}
\clearpage

\subsection{Expert-Level Complex Diagnostic}
\renewcommand{\arraystretch}{1.2}
\begin{figure}[!ht]
  \centering
  \begin{tabularx}{\textwidth}{@{} X @{}}
  \toprule
  \centering
  \includegraphics[width=7.5cm]{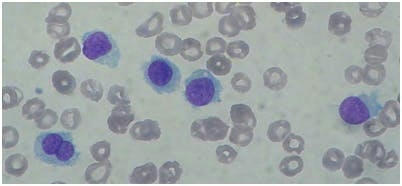}
  \vspace{1mm}

  \small
  \setlength{\parskip}{0.5ex}
  \begin{itemize}[leftmargin=0.5em, labelsep=0.5em]
    \item[] \textbf{Question} \par
    A 64-year-old female presents to her physician with complaints of persistent fatigue. She reports a decreased appetite, describing a constant sensation of fullness and lack of hunger. She also mentions frequent illnesses, noting this is her fourth clinic visit in six months. Her vital signs are stable, but she appears pale and distressed. Physical examination reveals splenomegaly as the only significant finding. Blood tests demonstrate pancytopenia, and her peripheral smear is shown below. Which of the following characteristics is associated with the cells observed on her peripheral smear?    
    
    Answer Choices: (A) Tartrate-resistant acid phosphatase positive. (B) Myeloid cells in all stages of development. (C) Myeloperoxidase positive Auer rods. (D) > 20\% lymphoblast cells in the peripheral smear.
    
    \item[] \textbf{Model Response}\par
    \texttt{<think>} 
    Okay, let's try to figure out this question. First, the question is about a 64-year-old female with fatigue, fullness, frequent illnesses, and splenomegaly. The labs show pancytopenia, and we need to look at the peripheral smear to identify a characteristic of the cells shown.

    First, let's look at the smear. The image shows several large, abnormal lymphoid cells. They have a fair amount of pale, slightly basophilic cytoplasm. The key feature is the cell border – it's not smooth. It has these fine, irregular, hair-like projections all around it. The nuclei are round to oval, and the chromatin looks a bit loose or stippled, not densely clumped. These look like classic "hairy cells".

    Now, let's connect this to the clinical picture. An older adult with pancytopenia (explaining fatigue/pallor, infections, and likely bleeding risk), splenomegaly (causing the sensation of fullness), and these "hairy cells" on the smear is the textbook presentation of Hairy Cell Leukemia (HCL). So, the question is really asking: which of these options is a characteristic of Hairy Cell Leukemia?
    
    Let's think about each option.
    
    Option A: Tartrate-resistant acid phosphatase positive (TRAP+). I remember that HCL is famous for this. The hairy cells are positive for acid phosphatase, and this activity is not inhibited by tartrate. This was a classic diagnostic test before flow cytometry became standard. This seems very likely to be correct.
    
    Option B: Myeloid cells in all stages of development. This describes the peripheral smear of Chronic Myeloid Leukemia (CML). CML can cause splenomegaly, but the cells are granulocytes (neutrophils, basophils, etc.), not these lymphoid cells with hairy projections. So, this is incorrect.
    
    Option C: Myeloperoxidase positive Auer rods. Auer rods are needle-like inclusions found in myeloblasts. This is a hallmark of Acute Myeloid Leukemia (AML). The cells in the smear are not blasts, and they don't have Auer rods. So, this is incorrect.
    
    Option D: > 20\% lymphoblast cells in the peripheral smear. This is the definition of Acute Lymphoblastic Leukemia (ALL). Lymphoblasts are typically smaller, with very scant cytoplasm, a high nuclear-to-cytoplasmic ratio, and condensed chromatin. The cells in the smear are larger and have that distinct "hairy" cytoplasm. So, this is incorrect.
    
    So, the patient's clinical presentation and the peripheral smear are classic for Hairy Cell Leukemia. The most specific and well-known characteristic of hairy cells among the choices is that they are TRAP positive.
    
    Yes, that makes sense. So the answer is A.
    
  \texttt{</think>}\par
   \item[] \textbf{Answer}: A.
   \end{itemize} \tabularnewline
  \bottomrule
  \end{tabularx}
\caption{An example of a complex diagnostic. The model links pancytopenia with splenomegaly to “hairy” lymphoid cells on the peripheral smear and concludes Hairy Cell Leukemia (TRAP+).}
\label{fig:expert_diag}
\end{figure}
\clearpage

\subsection{Real-World Clinical Report Generation}
\label{sec:rk_taoge_stage2_case_1of3}
\begin{figure}[H] 
  \centering
  \begin{tabularx}{\textwidth}{@{} X @{}}
  \toprule
  
  {
    \centering
    \includegraphics[height=3.3cm, keepaspectratio]{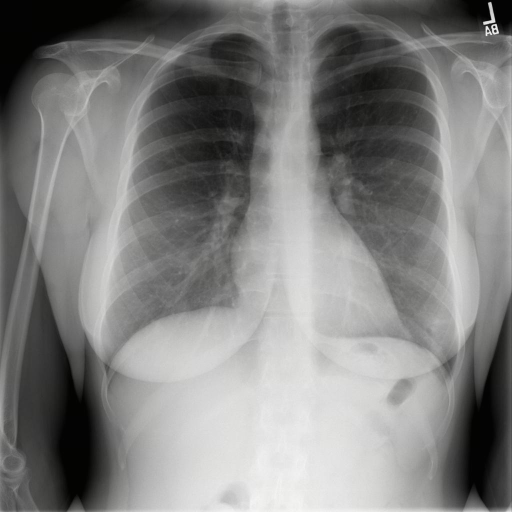}
    \hspace{0.05cm} 
    \includegraphics[height=3.3cm, keepaspectratio]{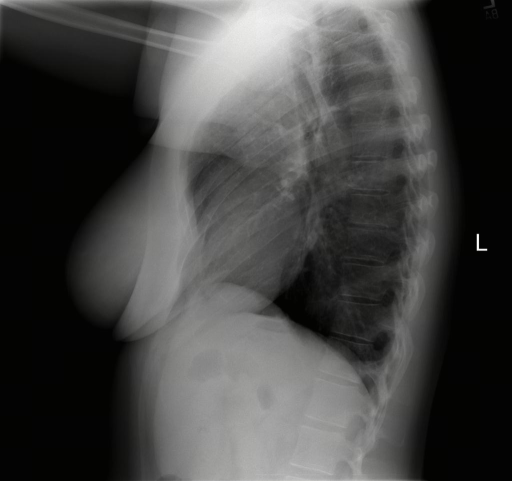}
    \hspace{0.05cm} 
    \includegraphics[height=3.3cm, keepaspectratio]{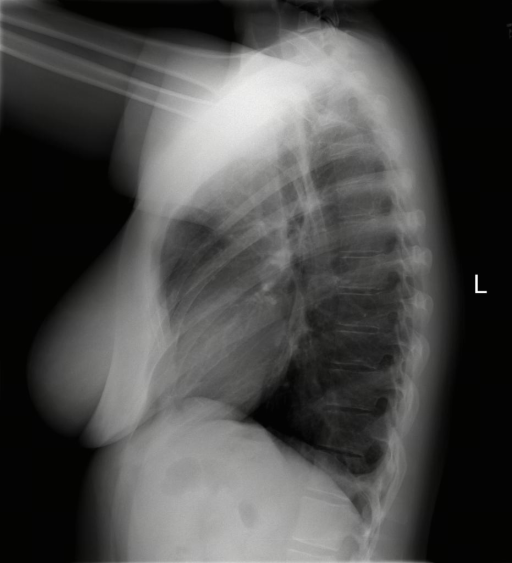}
  }
  
  \vspace{1em}
  
  \small
  \setlength{\parskip}{0.5ex}
  
  \textbf{Prompt:} You are a helpful assistant. Please generate a report for the given images, including both findings and impressions. Return the report in the following format: Findings:\{\}Impression:\{\}.
  
  \medskip 
  \textbf{Model Response}\par
  \begin{itemize}[leftmargin=4.5em, labelsep=0.5em]
      \item[\textcolor{teal}{\textbf{Step 1:}}]{\textbf{Task Understanding}}\newline
        I need to generate an IU - XRAY style report that includes Findings (objective observations from the chest X - rays) and Impression (clinical interpretation). The report must follow the specified format.

      \item[\textcolor{teal}{\textbf{Step 2:}}]{\textbf{Image Observation}}\newline
        Analyze key structures in each view:    - \textbf{Posteroanterior (PA) view}:    - Trachea: Appears as a longitudinal lucent band and is at the midline.    - Cardiomediastinal silhouette: Has a regular shape.    - Lungs: Mainly show black lucent areas; there are no focal high - density infiltrates, no blunting of costophrenic angles (so no pleural effusion), and no abnormal lucent space between the lung and chest wall (so no pneumothorax).    - Bony structures (clavicles, ribs, scapulae): The bone cortex is continuous, with no acute fractures or dislocations.  - \textbf{First lateral view}:    - Thoracic spine: Vertebral alignment is clear and the bone cortex is continuous.    - Lungs: The gray density is uniform, with no focal high - density opacities.    - Relationship between heart, thoracic spine, and sternum: No abnormal shift.    - Costophrenic angle: No increased density or shape change (so no effusion).    - Chest wall soft tissues and bony structures (ribs, thoracic spine): No signs of acute injury.  - \textbf{Second lateral view}:    - Thoracic spine and ribs: The bony structure is intact, with no fractures or osteolytic lesions.    - Lungs: The lucency is uniform, with no focal infiltration or abnormal lucent areas caused by pneumothorax.    - Mediastinal and cardiac contours: No widening or deformation.    - Diaphragm: The position and shape are normal, with no blurring of the diaphragm margin or elevation related to pleural effusion.

      \item[\textcolor{teal}{\textbf{Step 3:}}]{\textbf{Correlation \& Reasoning}}\newline
        Link observations to medical knowledge:  - Trachea at midline: Indicates no tracheal deviation, which could be caused by conditions like tension pneumothorax or large pleural effusion.  - Clear lungs without focal infiltrates: Rules out acute pneumonia, pulmonary edema, and other focal parenchymal diseases.  - No pleural effusion (sharp costophrenic angles, normal diaphragm): Excludes fluid accumulation in the pleural space.  - No pneumothorax (no abnormal pleural air): Rules out this acute and potentially life - threatening condition.  - Normal bony structures: Indicates no acute fractures or bony lesions, such as those from traumatic injury or osteomyelitis.  - Normal mediastinal, cardiac, and diaphragmatic findings: Shows no signs of cardiomegaly, mediastinal widening (e.g., from hematoma or adenopathy), or diaphragmatic dysfunction.
      \item[\textcolor{teal}{\textbf{Step 4:}}]{\textbf{Structuring the Report}}\newline
        For Findings, systematically report the status of each component (trachea, mediastinum, lungs, pleural spaces, bony structures) across all views, keeping the description concise. For Impression, synthesize the Findings to state the absence of acute cardiopulmonary or bony abnormalities.
  \end{itemize}

  \medskip 
  \textbf{Final Answer:} \newline
  Findings: The trachea is midline. The cardiomediastinal silhouette is normal. The lungs are clear, without evidence of focal infiltrate or effusion. There is no pneumothorax. There is no evidence of acute or chronic bone injury. \newline
  Impression: No obvious cardiopulmonary abnormalities were observed.
  \\
  \bottomrule
  \end{tabularx}
\caption{Example of radiology report generation with structured CoT, showing three input chest X-ray views, the prompt, stepwise reasoning, and the final Findings/Impression output.}
\label{fig:rk_taoge_stage2_case_1of3}
\end{figure}

\clearpage

\subsection{Deformed Clinical Report Understanding}

\begin{figure}[!ht]
  \centering
  \small 
  \renewcommand{\arraystretch}{1.2} 
  
  \begin{tabular}{p{0.96\linewidth}}
    \toprule
    
    \textbf{System:} You are an OCR QA model. Your task is to generate a response compliant with the requirements based on the input image and the user's question. The response requirements are as follows:
    \vspace{-5pt}
    \begin{itemize}\setlength{\itemsep}{0pt}
        \item The answer must be based on the image and cannot exceed the information provided.
        \item If the image contains no relevant information, the answer must directly be ``NOT\_FOUND''.
        \item Use the recognized information to generate the answer. Do not alter the recognized content unless required.
        \item Output the answer directly; do not include any explanation or additional text.
    \end{itemize}
    \vspace{5pt}
    
    \textbf{User Input:} \newline
    \vspace{2mm}
    {\centering \includegraphics[width=0.4\linewidth]{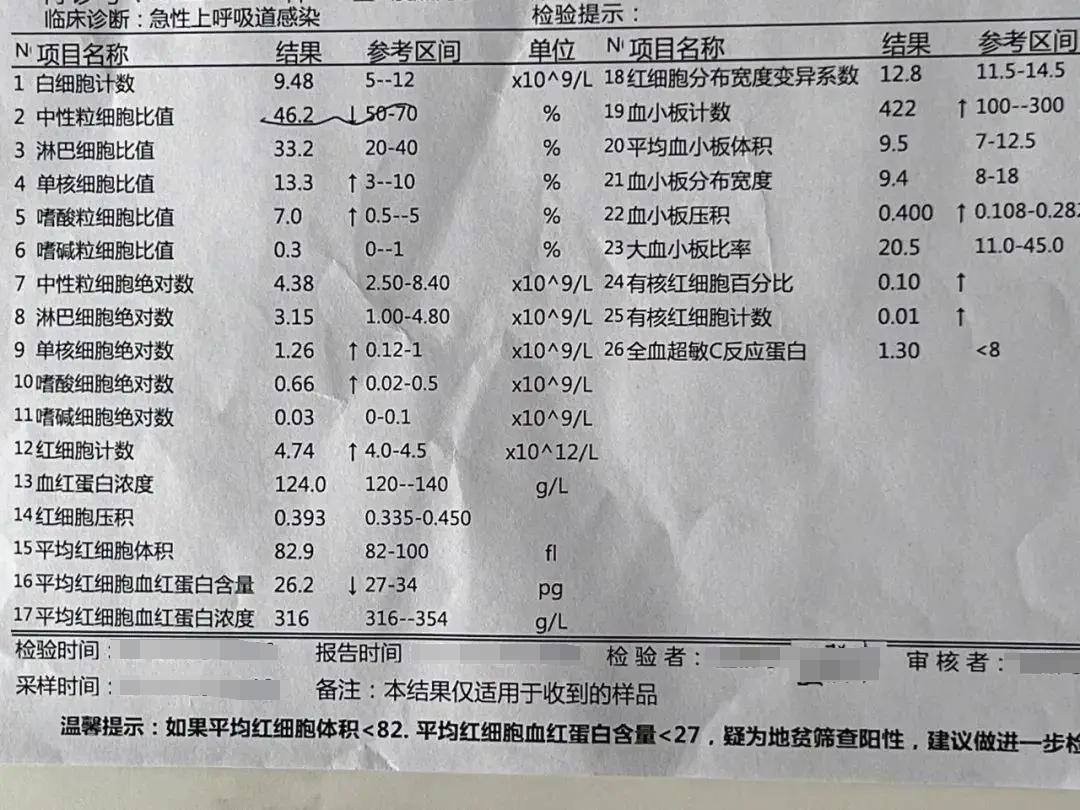} \par}
    \vspace{2mm}
    \textbf{User}: Is the measured value of mean corpuscular hemoglobin (MCH) within the reference range? \\[2mm]
    
    \midrule
    \addlinespace[2mm]
    
    \textbf{Model Responses:}\par
    
    \texttt{<think>} The user now needs to judge whether the measured value of mean corpuscular hemoglobin (MCH) is within the reference range. Based on the analysis, I should first use the ZOOM tool to magnify the area where this item appears. From the image, mean corpuscular hemoglobin is item 16, located roughly in the middle-lower part. First confirm imgidx is 0 (the first image). The label can be set to ``Mean Corpuscular Hemoglobin''. The bbox\_str should cover the row; assume coordinates are \texttt{<bbox>10 700 990 750</bbox>}. Set scale to 2 for clearer viewing. So the first step is to call the ZOOM tool with these parameters. \texttt{</think>}\par
    
    \vspace{2mm}
    
    \noindent\texttt{<tool\_call>} 
    ``arguments": `` \{``label":``Mean Corpuscular Hemoglobin",``imgidx":0,``bbox\_str":``<bbox>10 700 990 750</bbox>",``scale":2\}", ``name": ``ZOOM"
    \texttt{</tool\_call>}\par
    
    \vspace{2mm}
    
    {\centering \includegraphics[width=0.8\linewidth]{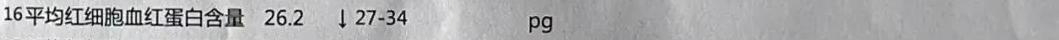} \par}
    
    \vspace{2mm}
    
    \texttt{<think>} The zoomed result shows the mean corpuscular hemoglobin (MCH) measured value is 26.2, and the reference interval is 27--34. A downward arrow indicates it is below the reference range. Therefore, the measured value is not within the reference range. No further tool calls are needed. \texttt{</think>}\par
    
    \vspace{2mm}
    
    \textbf{Answer}: The MCH measured value is 26.2 pg and the reference range is 27--34 pg, so it is below the reference range and not within it. \\
    \bottomrule
  \end{tabular}
  
  \caption{An example of precise information extraction from a deformed clinical report. The model identifies the specific row in the measurement table, uses the ZOOM tool to correct for resolution and deformation issues, and accurately extracts the numerical value requested by the user. The above content was originally in Chinese.}
  \label{fig:heart_report_qa}
\end{figure}

\clearpage

\section{Evaluation Details}
\label{app:benchmark_details}
\label{app:eval_detail}

\begin{table}[ht]
\centering
\caption{Public benchmarks for \textbf{Visual Diagnosis \& Image Recognition}.}
\label{tab:public_visual_diagnosis}
\footnotesize
\setlength{\tabcolsep}{7pt}
\renewcommand{\arraystretch}{1.12}
\begin{tabularx}{\columnwidth}{@{}l >{\raggedright\arraybackslash}X@{}}
\toprule
\textbf{Benchmark} & \textbf{What it is designed to evaluate} \\
\midrule
MMMU\_val-Med\citep{mmmu2024} &
Broad medical multimodal understanding and reasoning across disciplines; tests whether models connect visual evidence with domain knowledge. \\
MMMU\_Pro-Med\citep{mmmupro2025} &
Harder medical subset emphasizing advanced perception and complex reasoning under stronger distractors and tighter requirements. \\
\bottomrule
\end{tabularx}
\end{table}

\begin{table}[ht]
\centering
\caption{Public benchmarks for \textbf{Medical Imaging}.}
\label{tab:public_medical_imaging}
\footnotesize
\setlength{\tabcolsep}{7pt}
\renewcommand{\arraystretch}{1.12}
\begin{tabularx}{\columnwidth}{@{}l >{\raggedright\arraybackslash}X@{}}
\toprule
\textbf{Benchmark} & \textbf{What it is designed to evaluate} \\
\midrule
GMAI-MMBench\citep{gmaimmbench2024} &
Broad medical visual evaluation across many modalities and clinical tasks; emphasizes coverage and robustness. \\
VQA\_RAD\citep{vqarad2018} &
Radiology VQA requiring clinically grounded visual understanding and logical reasoning over imaging evidence. \\
SLAKE\citep{slake2021} &
Knowledge-enhanced medical VQA; probes multimodal fusion and structured medical knowledge usage. \\
PATH\_VQA\citep{pathvqa2020} &
Pathology microscopy understanding with textbook-level knowledge; stresses fine-grained morphology and interpretation. \\
PMC\_VQA\citep{pmcvqa2023} &
Generative VQA from biomedical literature figures and captions; tests figure understanding and biomedical grounding. \\
OmniMedVQA\citep{Hu2024OmniMedVQA} &
Large-scale unified VQA spanning multiple modalities and anatomy; stresses long-tail coverage and generalization. \\
\bottomrule
\end{tabularx}
\end{table}

\begin{table}[ht]
\centering
\caption{Public benchmarks for \textbf{Diagnosis}}
\label{tab:public_diagnosis}
\footnotesize
\setlength{\tabcolsep}{7pt}
\renewcommand{\arraystretch}{1.12}
\begin{tabularx}{\columnwidth}{@{}l >{\raggedright\arraybackslash}X@{}}
\toprule
\textbf{Benchmark} & \textbf{What it is designed to evaluate} \\
\midrule
MedXpertQA-MM\citep{Zuo2025MedXpertQA} &
Expert-level multimodal medical questions requiring integrated clinical reasoning beyond single-modality cues. \\
MedXpertQA-Text\citep{Zuo2025MedXpertQA} &
Expert-level text questions emphasizing deep medical knowledge reasoning and robust clinical judgement. \\
DiagnosisArena\citep{Zhu2025DiagnosisArena} &
Case-based diagnostic reasoning from high-quality sources; evaluates end-to-end evidence synthesis and conclusions. \\
HLE-Med\citep{phan2025humanity} &
Expert-challenging closed-form problems; probes near-ceiling reasoning and failure modes under difficulty spikes. \\
RareBench\citep{Chen2024RareBench} &
Rare disease reasoning, phenotype extraction, and differential diagnosis; stresses long-tail medical coverage. \\
MedBrowseComp\citep{medbrowsecomp2025} &
Medical deep-search / browsing agents; evaluates evidence gathering and fact synthesis for decision making. \\
\bottomrule
\end{tabularx}
\end{table}

\clearpage
\begin{table}[t]
\centering
\caption{Public benchmarks for \textbf{Medical Text}.}
\label{tab:public_medical_text}
\footnotesize
\setlength{\tabcolsep}{7pt}
\renewcommand{\arraystretch}{1.12}
\begin{tabularx}{\columnwidth}{@{}l >{\raggedright\arraybackslash}X@{}}
\toprule
\textbf{Benchmark} & \textbf{What it is designed to evaluate} \\
\midrule
PubMedQA\citep{pubmedqa2019} &
Reasoning over biomedical abstracts (background, results, conclusions); tests evidence-based reading comprehension. \\
MedQA\_MCMLE\citep{pal2022medmcqa} &
Chinese physician qualification-style questions; probes professional knowledge and clinical reasoning. \\
MedQA\_USMLE\citep{medqa2020} &
US medical licensing-style questions; evaluates English medical knowledge and clinical reasoning. \\
Medbullets\_op4\citep{ChenFangSinglaDredze2025Medbullets} &
Clinical MCQ recall and application with 4 options; emphasizes accurate discrimination under limited choices. \\
Medbullets\_op5\citep{ChenFangSinglaDredze2025Medbullets} &
Harder variant with more distractors; stresses robustness against confounders. \\
SuperGPQA-Med\citep{pteam2025supergpqascalingllmevaluation} &
Graduate-level interdisciplinary medical questions; evaluates depth and compositional reasoning. \\
CMExam\citep{cmexam2023} &
Large-scale Chinese medical exam benchmark; tests coverage and reasoning with fine-grained metadata. \\
MedMCQA\citep{pal2022medmcqa} &
Broad-coverage medical MCQs; evaluates general medical knowledge across subjects and difficulty levels. \\
HealthBench-hard\citep{healthbench2025} &
Difficult multi-turn medical dialogues; stresses correctness, safety, and communication quality under hard scenarios. \\
HealthBench-normal\citep{healthbench2025} &
Realistic medical conversations; evaluates helpfulness, alignment, and safety in typical patient-facing settings. \\
\bottomrule
\end{tabularx}
\end{table}

\begin{table}[ht]
\centering
\caption{Public benchmarks for \textbf{Medical Report}.}
\label{tab:public_medical_report}
\footnotesize
\setlength{\tabcolsep}{7pt}
\renewcommand{\arraystretch}{1.12}
\begin{tabularx}{\columnwidth}{@{}l >{\raggedright\arraybackslash}X@{}}
\toprule
\textbf{Benchmark} & \textbf{What it is designed to evaluate} \\
\midrule
MIMIC-CXR\citep{mimiccxr2019} &
Chest X-ray understanding and report generation; tests image-text alignment and clinically grounded descriptions. \\
CheXpert Plus\citep{chexpertplus2024} &
Fine-grained chest X-ray understanding; probes pathology-level discrimination beyond coarse labels. \\
IU-Xray\citep{iuxray2016} &
Radiology report/caption generation; evaluates descriptive faithfulness and content coverage. \\
\bottomrule
\end{tabularx}
\end{table}

\begin{table}[ht]
\centering
\caption{Public benchmarks for \textbf{Instruction Following}.}
\label{tab:public_instruction_following}
\footnotesize
\setlength{\tabcolsep}{7pt}
\renewcommand{\arraystretch}{1.12}
\begin{tabularx}{\columnwidth}{@{}l >{\raggedright\arraybackslash}X@{}}
\toprule
\textbf{Benchmark} & \textbf{What it is designed to evaluate} \\
\midrule
MultiChallenge\citep{multichallenge2024} &
Complex multi-turn medical dialogues; tests constraint tracking, consistency, and dialogue memory. \\
MedMTbench\citep{yang2025medmtbench} &
Long-context medical instruction following with implicit demands; stresses real-world conversational controllability. \\
MulDimIF\citep{MulDimIF} &
Constraint-based instruction following across multiple patterns, categories (e.g., content/language/format/length), and difficulty levels using code-verifiable checks. \\
\bottomrule
\end{tabularx}
\end{table}

\clearpage
\section{Prompts for Entity Extraction}
\label{app:app_prompts_for_entity_extraction}

\begin{center}
\begin{tcolorbox}[colback=myLightCyan, colframe=myTeal, title=\textbf{Prompt for Stage 2: Joint Extraction and Typing}]
\small
\texttt{
Below are several sentences. Analyze these sentences for **medical** entity nouns and their types, and output according to the following requirements:
\begin{enumerate}
    \item Entity nouns must be informative proper nouns. Secondarily, determine if they are **medical** entities; if not, do not output.
    \item Pay attention to overly long medical entity nouns and determine if they can be segmented/split.
    \item The sentences below may contain special symbols and meaningless spaces; please ignore them directly.
    \item Replace \textless EntityType\textgreater\ with the specific entity category.
    \item Replace \textless EntityName\textgreater\ with the specific entity noun.
    \item Output strictly in JSON format. The example format is as follows: \\
    \textbf{\{\{
        'Sentence0': [\textless EntityType\textgreater:\textless EntityName\textgreater, ...], 
        'Sentence1': [...], 
        ...
    \}\}}
\end{enumerate}
Sentences:\\
\{lines\}
}
\label{app:prompts_joint_extra}
\end{tcolorbox}
\end{center}

\begin{center}
\begin{tcolorbox}[colback=myLightCyan, colframe=myTeal, title=\textbf{Prompt for Stage 3: Entity Tree Attachment}]
\small
You are a medical entity taxonomy expert. You must strictly follow the guidelines below to integrate the list of medical entity nouns I provide into the existing medical entity tree.

\textbf{**Maintain Original Structure**}:\\
You are NOT allowed to change the names of nodes on the medical entity tree.

\textbf{**Precise Insertion**}:\\
Insert the medical entity nouns into the appropriate sub-categories of their parent nodes. Ensure only valid medical entities are inserted and adhere to hierarchical relationships.

\textbf{**Rationality of New Insertions**}:\\
If you insert a medical entity into the tree, you must provide the reason for its insertion to ensure interpretability and traceability.\\
Output Format: \textless Reason\textgreater xxx\textless /Reason\textgreater \textless InsertionPath\textgreater Node1.Node2.InsertedNode\textless /InsertionPath\textgreater

\textbf{**Handling Unclassifiable Cases**}:\\
Report medical entity nouns that cannot be classified and explain the reasons for the uncertainty. If unsure about the classification, provide detailed reasoning.\\
Output Format: \textless Reason\textgreater xxx\textless /Reason\textgreater \textless Reasoning\textgreater yyy\textless /Reasoning\textgreater

\vspace{0.2cm}
Medical Entity Tree:\\
\{tree\}

Medical Entity Nouns:\\
\{entity\}
\label{app:prompts_entity_tree_attch}
\end{tcolorbox}
\end{center}

\begin{center}
\begin{tcolorbox}[colback=myLightCyan, colframe=myTeal, title=\textbf{Prompt for Stage 3 Agent: Conflict Resolution}]
\small
You are a rigorous medical knowledge base construction Agent. Your task is to resolve entity classification conflicts.
The current entity "\{entity\}" is attached to multiple parent nodes:
1. Parent Path A: \{path\_a\}
2. Parent Path B: \{path\_b\}
...

You must search for the exact medical definition via Google/Wiki and adjudicate based on the following principles:
\begin{enumerate}
    \item \textbf{Principle of Etiological Dominance}: Classification based on pathological mechanism or anatomical location takes precedence over clinical symptoms.
    \item \textbf{Principle of Specificity}: If one parent is a subset of another and describes the entity more accurately, prefer the more specific one.
\end{enumerate}

Thinking Steps:
Step 1: Construct query terms and call tools to search for the definition.
Step 2: Analyze results and compare the validity of each parent path.
Step 3: Decide which path to keep and delete the others.

Output Format:
\textless SearchEvidence\textgreater Excerpt from Wiki/Search...\textless /SearchEvidence\textgreater
\textless Reasoning\textgreater Since evidence shows..., and Path A focuses on XX while Path B focuses on XX, according to the Principle of Etiological Dominance, Path A is more appropriate.\textless /Reasoning\textgreater
\textless FinalAction\textgreater Keep: \{path\_a\}, Delete: \{path\_b\}\textless /FinalAction\textgreater
\label{app:prompts_entity_conflict_enhance}
\end{tcolorbox}
\end{center}

\end{document}